\documentclass[letterpaper, 10 pt, journal, twoside]{IEEEtran}

\PassOptionsToPackage{table}{xcolor}
\usepackage{xcolor}

\usepackage[T1]{fontenc}
\usepackage{cite}
\usepackage{amssymb,amsfonts}
\usepackage{blindtext}
\makeatletter
\let\NAT@parse\undefined
\makeatother
\usepackage{hyperref}
\usepackage{algorithmic}
\usepackage{graphicx}
\usepackage{textcomp}
\usepackage{mathtools}
\usepackage{changes}
\usepackage[font=footnotesize,labelformat=simple]{subcaption}
\usepackage{algorithm}
\usepackage{color, soul}
\usepackage{amsmath}
\usepackage{booktabs}
\usepackage{multirow}
\usepackage{xstring}
\usepackage{hhline}
\usepackage{kotex}
\usepackage{siunitx}
\usepackage{comment}
\usepackage{physics}
\usepackage{tikz}
\usepackage{cleveref}
\usepackage{gensymb}
\usepackage{threeparttable}
\usepackage{caption}
\usepackage{url_bib_utils}

\usepackage{mathtools}
\usepackage{soul}
\usepackage{amssymb}
\usepackage{amsfonts}
\usepackage{pifont}
\usepackage{multirow}

\crefformat{section}{\S#2#1#3} 
\crefformat{subsection}{\S#2#1#3}
\crefformat{subsubsection}{\S#2#1#3}

\def\eqref#1{Eq.~(\ref{#1})}

\def\figinitref#1{Fig.~\ref{#1}}
\DeclareMathOperator*{\argmax}{arg\,max}

\newcommand{\concat}{\mathbin{\oplus}}

\newcommand*{\mvonelink}{\url{https://youtu.be/DECFbMdpfps}}
\newcommand*{\mvtwolink}{\url{https://youtu.be/lAdPj3KTDo8}}
\newcommand*{\mvthreelink}{\url{https://youtu.be/mgNLLNxg52A}}
\newcommand*{\mvfourlink}{\url{https://youtu.be/G_lTWikijWk}}
\newcommand*{\mvfivelink}{\url{https://youtu.be/tFI8xDwF4bU}}
\newcommand*{\mvsixlink}{\url{https://youtu.be/lESpqB5LTnI}}
\newcommand*{\mvsevenlink}{\url{https://youtu.be/x3RSJRUr7ro}}
\newcommand*{\mveightlink}{\url{https://youtu.be/pwwwmnd3Xnc}}

\newcommand*{\mvtenlink}{\url{https://youtu.be/XVC7c5DlB4I}}
\newcommand*{\mvonefoot}{\footnote{\mvonelink}}
\newcommand*{\mvtwofoot}{\footnote{\mvtwolink}}
\newcommand*{\mvthreefoot}{\footnote{\mvthreelink}}
\newcommand*{\mvfourfoot}{\footnote{\mvfourlink}}
\newcommand*{\mvfivefoot}{\footnote{\mvfivelink}}
\newcommand*{\mvsixfoot}{\footnote{\mvsixlink}}
\newcommand*{\mvsevenfoot}{\footnote{\mvsevenlink}}
\newcommand*{\mveightfoot}{\footnote{\mveightlink}}

\newcommand*{\mvtenfoot}{\footnote{\mvtenlink}}

\definecolor{best}{RGB}{168, 192, 251}     
\definecolor{ab_better}{RGB}{193, 210, 251}   
\definecolor{second}{RGB}{229, 234, 251}     
\definecolor{ab_bad}{RGB}{251, 233, 233}
\definecolor{ab_worse}{RGB}{245, 201, 201}
\definecolor{worst}{RGB}{239, 169, 169}
\newcommand{\hlbest}[1]{\sethlcolor{best}\hl{#1}}
\newcommand{\hlsecond}[1]{\sethlcolor{second}\hl{#1}}
\newcommand{\hlworst}[1]{\sethlcolor{worst}\hl{#1}}
\definecolor{rev}{RGB}{0, 0, 255}

\newcommand{\alignedcell}[2]{\cellcolor{#1}\raisebox{-0.2ex}{$#2$}}

\definecolor{tbd}{RGB}{255, 0, 0}
\definecolor{robot_R1}{RGB}{0,180,0}
\definecolor{robot_R2}{RGB}{0,180,180}
\definecolor{robot_R3}{RGB}{225,0,179}
\definecolor{instructioncolor}{rgb}{0., 0.0, 0.}
\definecolor{add}{rgb}{0, 0, 255}

\renewcommand{\thesubsection}{\thesection.\Alph{subsection}}

\IEEEoverridecommandlockouts                              

\title{DreamWaQ++: Obstacle-Aware Quadrupedal Locomotion With Resilient Multi-Modal Reinforcement Learning}

\author{I Made Aswin Nahrendra$^{1,2\dag}$, Byeongho Yu$^{3}$, Minho Oh$^{1,3}$, Dongkyu Lee$^{1,3}$, \\
Seunghyun Lee$^{1}$, Hyeonwoo Lee$^{1}$, Hyungtae Lim$^{4}$, Hyun Myung$^{1\ast}$%
\thanks{$^1$Urban Robotics Lab., School of Electrical Engineering, KAIST, Daejeon, Republic of Korea.}%
\thanks{$^2$KRAFTON, Seoul, Republic of Korea.}%
\thanks{$^3$URobotics, Seoul, Republic of Korea.}%
\thanks{$^4$Laboratory for Information and Decision Systems (LIDS), MIT, Cambridge, MA, USA.}%
\thanks{$^{\dag}$Currently at $^2$. This work was done during his time at $^1$.}%
\thanks{$^\ast$Corresponding author: Hyun Myung ({\tt hmyung@kaist.ac.kr})}%
\thanks{Project page: \url{https://dreamwaqpp.github.io}}%
\thanks{This work was supported in part by Korea Evaluation Institute of Industrial Technology (KEIT) funded by the Korea Government (MOTIE) under grant No. 20018216, ``Development of Mobile Intelligence SW for Autonomous Navigation of Legged Robots in Dynamic and Atypical Environments for Real Application'', and in part by the R\&D Program for Forest Science Technology (Project No. RS-2025-25424472) provided by Korea Forest Service (Korea Forestry Promotion Institute). The students are supported by BK21 FOUR.}%
}

\begin{document}


\maketitle

\IEEEpeerreviewmaketitle

\begin{abstract}
    Quadrupedal robots hold promising potential for applications in navigating cluttered environments with resilience akin to their animal counterparts. However, their floating-base configuration makes them susceptible to real-world uncertainties, presenting substantial challenges in locomotion control. Deep reinforcement learning has emerged as a viable alternative for developing robust locomotion controllers. However, approaches relying solely on proprioception often sacrifice collision-free locomotion, as they require front-foot contact to detect stairs and adapt the gait. Meanwhile, incorporating exteroception necessitates a precisely modeled map observed by exteroceptive sensors over time. This work proposes a novel method for fusing proprioception and exteroception through a resilient multi-modal reinforcement learning framework. The proposed method yields a controller demonstrating agile locomotion on a quadrupedal robot across diverse real-world courses, including rough terrains, steep slopes, and high-rise stairs, while maintaining robustness in out-of-distribution situations.
\end{abstract}


\section{Introduction}
\noindent In the past decade, quadrupedal robots have revolutionized robotic applications in real-world environments thanks to their capability to traverse cluttered spaces, enabling diverse applications spanning exploration and inspection~\cite{gehring2021anymal,tranzatto2022cerberus,hong2022agile,arm2023scientific}. The growing interest in quadrupedal robot applications has been accompanied by advancements in control algorithms, which have evolved from traditional model-based control~\cite{jenelten2022tamols,bellicoso2017dynamic,gehring2017quadrupedal,bledt2018cheetah,hong2020real} to data-driven approaches such as deep reinforcement learning (RL)~\cite{lee2020learning,kumar2021rma,ji2022concurrent,nahrendra2023dreamwaq,choi2023learning,miki2022learning,gangapurwala2022rloc,yang2022learning,imai2022vision,agarwal2022legged,yang2023neural}.

Traditional model-based control pipelines for legged robots typically rely on a complex cascaded structure~\cite{jenelten2022tamols} that comprises accurate state estimation~\cite{bloesch2013state,camurri2020pronto,kim2021legged,kim2022step}, terrain mapping~\cite{fankhauser2018probabilistic,lim2021patchwork,lee2022patchwork++,oh2022travel,miki2022elevation}, and a whole-body controller optimizing the robot's foot trajectory~\cite{bellicoso2017dynamic,gehring2017quadrupedal,bledt2018cheetah,hong2020real}. However, these pipelines can be computationally intensive for real-time inference and often require strict assumptions, such as collision-free and non-slip conditions. Although simplified models are sometimes used to reduce problem complexity, they potentially aggravate the performance.

\begin{figure}[t!]
    \centering 
    \begin{subfigure}[b]{0.48\textwidth}
        \includegraphics[width=1.0\textwidth]{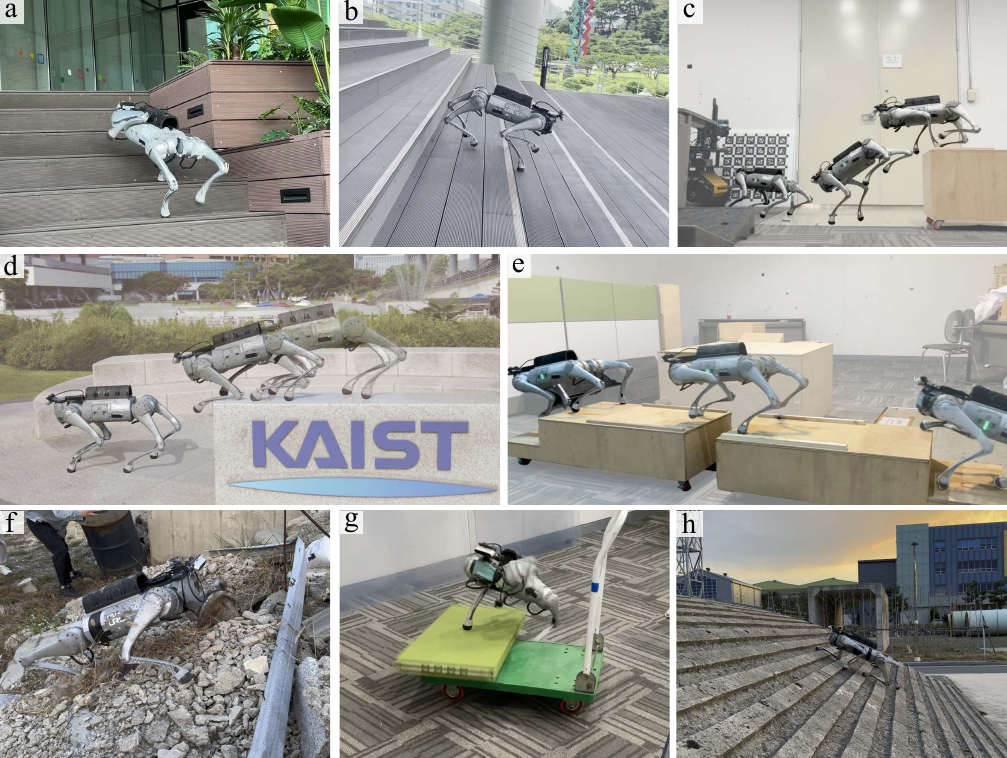}
    \end{subfigure}
    \caption{The locomotion controller trained using DreamWaQ++ allows a quadrupedal robot to perform agile and resilient locomotion over various obstacles and terrains. The controller exhibits versatile gaits such as (a)~ascending and (b)~descending over a flight of stairs, (c)~performing a leap motion, (d)~probing when faced with an uncertain dip, (e)~crossing a gap, (f)~adapting to unseen deformable disastrous terrain, (g)~balancing on movable platforms, and~(h)~climbing a $35^\circ$ slope. Note that all these behaviors are embodied in a single neural network without specialized training for a particular scenario.}
    \label{figure:dreamwaqpp_main_fig}
\end{figure}

As opposed to model-based control, deep RL methods reformulate the optimization problem into offline optimization during training by learning a decision-making policy that implicitly plans future control actions based on given observations. Notably, a blind locomotion controller, relying only on proprioception, demonstrates impressive robustness across various terrain profiles~\cite{lee2020learning,kumar2021rma,ji2022concurrent,nahrendra2023dreamwaq,choi2023learning}. However, the resilience of blind locomotion controllers is inherently limited, as they require collisions between the robot's legs and the environment to sense obstacle properties and adjust the gait accordingly.

To advance blind locomotion controllers, an efficient fusion of proprioception and exteroception to learn a robust quadrupedal locomotion controller is actively studied in the legged robotics community~\cite{miki2022learning,gangapurwala2022rloc,yang2022learning,imai2022vision,agarwal2022legged,yang2023neural}. Naturally, animals have an agile locomotion behavior, owing to their ability to observe the terrain ahead using their eyes and quickly plan their effective gait for traversing the terrain. Therefore, incorporating exteroception for the gait planning of legged robots is paramount for eliciting agile behaviors~\cite{ho2022people,di2019walking,chopra2018cognitively,killeen2017minimum}.

We propose DreamWaQ++, an obstacle-aware quadrupedal locomotion controller that specifically aims to tackle the following challenges: 1) a resilient controller with multi-modal perception capability and sensor-agnostic nature that can be integrated with various options of exteroceptive sensors, 2) an efficient control framework that enables real-time control and fast adaptation, 3) an efficient reinforcement learning (RL) pipeline with a single-stage learning procedure. By employing DreamWaQ++ on a Unitree Go1~\cite{unitreego1}, we demonstrated remarkable performance in traversing various challenging environments\mvonefoot as shown in \figinitref{figure:dreamwaqpp_main_fig}.

This paper is an evolved version of the conference paper~\cite{nahrendra2023dreamwaq}. We have extended the work by providing substantial improvement in the controller's performance and robustness in the following aspects:
\begin{itemize}
    \item Incorporation of a novel multi-modal perception module that fuses proprioception and exteroception. The module enables the controller to adapt to various terrains and obstacles, including stairs, gaps, and deformable terrains.
    \item Enhanced robustness via a multi-modal deep RL framework that enables the controller to efficiently leverage the multi-modal perception module for learning a robust locomotion policy in a single-stage learning procedure.
    \item Improved agility via skill discovery objectives that encourage the controller to learn versatile gaits for traversing diverse terrains and obstacles.
    \item Extensive evaluation on various challenging environments and robotics platforms, highlighting the robustness and scalability of the proposed controller.
    \item Ablation studies to analyze the framework's components and their contributions to the controller's performance, opening up new research directions for future work.
\end{itemize}

\section{Related Works}
In recent years, research on legged locomotion has made significant progress, largely driven by advancements in simulation-based policy learning through deep reinforcement learning. Earlier works in the field primarily focused on blind locomotion strategies, aiming to ensure reliable sim-to-real transfer by minimizing actuator model discrepancies~\cite{hwangbo2019learning} and incorporating adaptation mechanisms to handle environmental variations~\cite{lee2020learning,kumar2021rma,ji2022concurrent,nahrendra2023dreamwaq}. These approaches relied predominantly on proprioceptive inputs to control the robot's movement. However, due to their limited perceptual capabilities, these methods often struggle in more complex dynamic environments.

To address these limitations, exteroceptive perception has been incorporated into the locomotion pipeline to enable policies that are not only stable but also aware of the surrounding terrain geometry, particularly near the robot’s legs~\cite{miki2022learning,gangapurwala2022rloc,agarwal2022legged,loquercio2023learning}. Recent studies have utilized raw egocentric depth vision for locomotion~\cite{yang2022learning,imai2022vision,agarwal2022legged,yang2023neural,cheng2023parkour}to mimic the locomotion abilities of animals. However, elevation map–based approaches~\cite{miki2022learning,gangapurwala2022rloc,hoeller2023anymal} have proven to be superior, particularly in situations where depth vision is unreliable due to limited field of view~(FoV).

In addition to exteroception, memory-based architectures such as long short-term memory (LSTM) and gated recurrent units (GRU) have become primary components in the success of recent perceptive locomotion controllers~\cite{miki2022learning,margolis2022rapid,ji2022concurrent,choi2023learning} to mitigate partial observability. However, training recurrent network models often suffers from vanishing gradients due to the backpropagation through time (BPTT) mechanism~\cite{werbos1990backpropagation}. As a workaround, variants of convolutional neural network (CNN) architectures have been used as a viable option to handle sequential data~\cite{lee2020learning,kumar2021rma}. However, CNNs are prone to inductive bias, which assumes that neighboring data points are more likely to be related than others. This inductive bias hinders a neural network’s ability to freely learn the positional relationships between features in unstructured time-series data. More recently, attention-based sequence models, such as transformers~\cite{vaswani2017attention,melo2022transformers}, have shown potential as viable alternatives for constructing memory in locomotion tasks~\cite{yang2022learning}.

Utilizing large and expressive models such as transformers is effective for learning complex policies in high-dimensional action spaces~\cite{melo2022transformers}. However, these models typically require large amounts of training data and are more susceptible to real-time inference constraints due to their computational overhead. To effectively balance perception, memory, and fast reaction speed, we propose a multi-modal mixer architecture that combines the strengths of both lightweight MLPs and attention-based mechanisms. This architecture enables efficient processing of both proprioceptive and exteroceptive inputs, allowing the policy to learn a robust and adaptable locomotion strategy.

However, memory alone is sometimes insufficient for achieving resilient locomotion behavior if the learned latent representation obtained from memory does not account for explorative behavior that promotes skill discovery. Without an adequate skill discovery strategy, the latent representation may lead the policy to overfit to a limited range of behaviors, resulting in a conservative policy that struggles to adapt to diverse environmental changes~\cite{li2023versatile,rakelly2021mutual}. Therefore, we propose two regularization techniques that balance accurate perception learning while also promoting explorative behavior. 
\section{The Proposed Learning Framework}

\subsection{Overview}
    \begin{figure*}[t!]
        \centering 
        \begin{subfigure}[b]{0.98\textwidth}
            \includegraphics[width=1.0\textwidth]{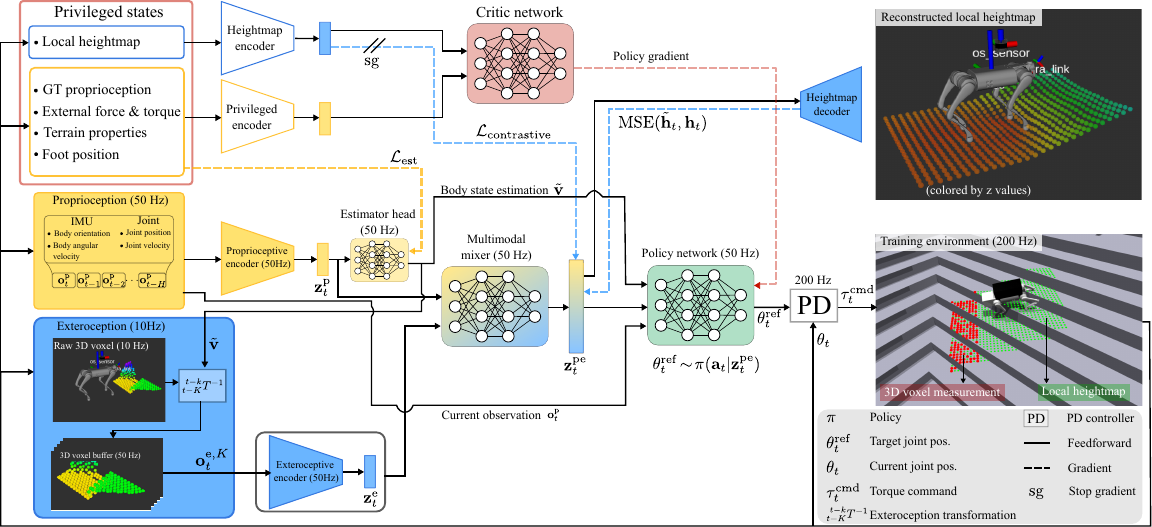}
        \end{subfigure}
        \caption{Overview of DreamWaQ++. The encoder has a hierarchical structure that consists of low-level raw measurement encoders and a spatio-temporal mixer.}
        \label{figure:dreamwaqpp}
    \end{figure*}
    DreamWaQ++ comprises a perception pipeline and a control pipeline, detailed throughout this section. All modules are jointly trained with a combination of objective functions that facilitate interaction between networks, promoting cooperative learning of informative latent features. This training approach resembles the context adaptation paradigm in a few-shot meta-RL setting~\cite{finn2017model,rakelly2019efficient}. Specifically, the context encoder is trained on a large dataset of simulated quadrupedal locomotion with extensive domain randomization, resembling the meta training phase. Subsequently, the learned context is used to condition the policy in real time, akin to the meta-testing phase. Simultaneously, the policy is trained to effectively control the robot, thereby mitigating the impact of poor context features and state estimation. All trained networks were deployed on a real-world quadrupedal robot with onboard sensors, without any fine-tuning. An overview of the entire framework is shown in Fig.~\ref{figure:dreamwaqpp}.

    The control problem is formulated in a partially observable Markov decision process~(POMDP) setting with a goal to maximize the expected discounted future rewards, which in turn results in a policy:

    \begin{equation}
        \pi=\argmax_{\textbf{a}} E\left[\sum_{t=0}^{\infty}\gamma^t r_t\right],
        \label{eq:pi}
    \end{equation}
    where $\textbf{a}$, $\gamma$, and $r$ are the action, discount factor, and rewards, respectively. This objective is optimized using the proximal policy optimization (PPO)~\cite{schulman2017proximal} algorithm, while also taking into account the auxiliary objectives related to the perception pipeline.

    We adopt a privileged learning setting using an asymmetric actor-critic architecture trained using the proximal policy optimization (PPO)~\cite{schulman2017proximal} algorithm. The actor, i.e. the policy receives partial and noisy observations ($\textbf{o}^\mathrm{p}_t$), akin to the real-world observations and the multi-modal context ($\textbf{z}^\mathrm{pe}_t$) as its input.  The critic, by contrast, receives a privileged state that are accessible only in simulation. The policy network runs at a rate of $50~\mathrm{Hz}$, generating a target joint position that is tracked by a low-level PD controller, running at $200~\mathrm{Hz}$, to generate the joint torque commands.

\subsection{Hierarchical Exteroceptive Memory}
Employing exteroceptive measurements in a controller presents distinct challenges due to the low-frequency nature of the sensor, which operates at approximately two to five times lower than the control loop and proprioceptive sensors. This asynchrony introduces non-negligible delays into the control loop, yielding degraded performance.

We circumvent this issue using a memory structure that generates a denser point cloud, $\textbf{o}^{\mathrm{e},K}_t$, around the robot by concatenating points from the last $K$ measurements, each transformed to the robot's current position using $SE(3)$ transformations. Unlike approaches such as~\cite{hoeller2022neural,hoeller2023anymal}, which employ U-Net-based architectures for full scene reconstruction, we use autoregression solely to estimate the $SE(3)$ transformation of the robot’s body frame over time. This method avoids computationally intensive reconstruction while still providing temporally dense exteroception for the controller.

The $SE(3)$ transformation is updated at each control loop iteration by combining the IMU's orientation measurement with the integration of the body linear velocity $\textbf{v}_t$ predicted by the state estimation network, which leverages the history of the robot’s proprioception as detailed in Section~~\ref{section:proprioceptive_encoder}. This transformation process enables temporal extrapolation of the latest exteroceptive measurements by aligning previous points with the robot’s current frame, forming the basis of a memory structure that is then fed into the exteroceptive encoder. The exteroceptive observation is formally defined as follows:

\begin{equation}
    \textbf{o}^{\mathrm{e},K}_t = \textbf{o}^\mathrm{e}_t \concat \hat{\textbf{o}}^\mathrm{e}_{t-1} \concat \cdots \concat \hat{\textbf{o}}^\mathrm{e}_{t-K}, 
  \end{equation}
   where $\textbf{o}^\mathrm{e}_t$ is the most recent exteroceptive observation at time $t$. $\hat{\textbf{o}}^\mathrm{e}_{t-K}$ is the previous exteroceptive observation at $t-K$, which has been transformed to the robot's body frame at time $t$, which is defined as
  \begin{equation}
    \hat{\textbf{o}}^\mathrm{e}_{t-k} = \prescript{t-k}{t-K}{T}^{-1}\cdot \textbf{o}^\mathrm{e}_{t-K},
  \end{equation}
where $\textbf{o}^\mathrm{e}_{t-k}$ is the exteroceptive observation measured at time $t-k$~$\left(k\in[0, K]\right)$, and $\prescript{t-k}{t-K}{T}$ is the $SE(3)$ transformation of the robot's pose from time $t$ to $t-k$. 

As the state estimation network only predicts the body linear velocity, we integrate this velocity over $k$ steps using Euler integration to estimate the robot's position transformation. Although drift may accumulate in the pose estimate over time, we found that this is not significant because the integration is reset every $K$ steps when a new exteroceptive measurement is received. Moreover, we also randomized the exteroceptive sensor pose with respect to the robot body frame during training and applied several levels of exteroception noise to enhance the policy's robustness against erroneous local pose estimation and exteroception data~(Section~\ref{section:adversarial}).

In this work, we set the exteroception sampling rate to $10~\textrm{Hz}$, resulting in $K=5$ for constructing the exteroceptive memory. While higher sampling rates are theoretically possible depending on the choice of sensor, we selected $10~\textrm{Hz}$ due to practical constraints. In particular, higher rates might be slightly unreliable due to data processing latency on the limited onboard computing resources. Moreover, using $10~\textrm{Hz}$ sampling rate also allows us to change the sensor into LiDAR without the need to retrain the policy, as commercial 3D LiDAR sensors typically operate at $10~\textrm{Hz}$.

\subsection{Exteroceptive Encoder}
\begin{figure}[t!]
	\centering 
	\begin{subfigure}[b]{0.48\textwidth}
		\includegraphics[width=1.0\textwidth]{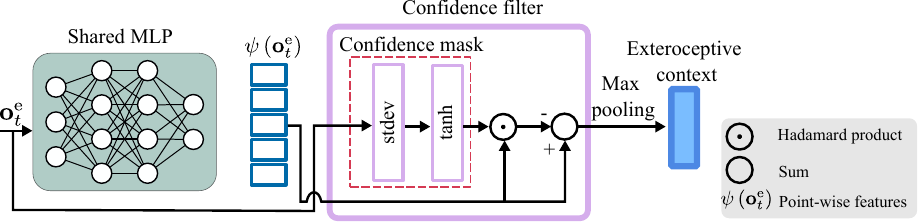}
	\end{subfigure}
	\caption{The exteroceptive encoder uses a PointNet-based architecture as its backbone. We designed a confidence filter layer that statistically learns a masking layer that cancels out unreliable point features before aggregating them into the exteroceptive context $\textbf{z}^\mathrm{e}_t$.}
	\label{figure:ext_encoder}
\end{figure}

We utilized 3D points as input to our framework to enable flexible compatibility with multiple sensor configurations, such as a 3D LiDAR sensor or depth camera. A PointNet-based architecture~\cite{qi2017pointnet} is employed to effectively extract information from the input point cloud, accommodating an arbitrary number of points while maintaining robustness to noise.

Although the max-pooling layers in PointNet enforce invariance to the number and order of input points in the point cloud, they can become detrimental when outliers and heavy noise dominate the input. This issue arises because the max-pooling operation aggregates point features indiscriminately. To address this, we employ a confidence filter layer following the backbone PointNet architecture (Fig.~\ref{figure:ext_encoder}). The confidence filter statistically rejects unreliable points in the latent space using a filtering operation, resulting in confidence-filtered points defined as:

\begin{equation}\label{eq:confidence_filter}
    \mathcal{C} \left( \textbf{o}^{\mathrm{e},K}_t \right) = \psi^\textrm{e} \left( \textbf{o}^{\mathrm{e},K}_t \right) \cdot\left( 1 - \tanh \left( \sigma \left( \textbf{o}^{\mathrm{e},K}_t\right) \right) \right),
\end{equation}
where $ \psi^\textrm{e} \left(\cdot\right)$ is the backbone PointNet layer, and $\sigma(\cdot)$ is a standard deviation operator that statistically assesses the diversity of the input point cloud. A hyperbolic tangent operation, $\tanh(\cdot)$, is used to smoothly set an upper bound of $\sigma \left( \textbf{o}^{\mathrm{e},K}_t \right)$ to one. Each point feature, $\psi\left(\textbf{o}^{\mathrm{e},K}_t\right)$, is fed into a shared confidence mask layer that outputs confidence masks based on the statistics of the raw points. The confidence mask outputs a value close to $1$ for high-variance features and a value close to $0$ for low-variance features, due to the $\tanh$ layer. Following \eqref{eq:confidence_filter}, $\mathcal{C} \left( \textbf{o}^{\textrm{e},K}_t \right)$ removes high-variance features, such as outliers, while preserving low-variance features. The filtered point features are then aggregated using max-pooling to obtain the exteroceptive context $\textbf{z}^\mathrm{e}_t$.

\subsection{Proprioceptive Encoder}\label{section:proprioceptive_encoder}
The proprioceptive encoder is based on the context-aided estimator network (CENet)~\cite{nahrendra2023dreamwaq}, modified by replacing standard fully connected layers with an MLP-mixer architecture~\cite{tolstikhin2021mlp}. This modification enables interactions across different proprioceptive modalities over multiple time frames, enhancing both the explicit estimation and latent representation of proprioception. To achieve this, we treat the proprioceptive features as tokens and the time frames as channels, enabling the MLP-mixer to learn interactions across different proprioceptive modalities and temporal dimensions. We employ MLPs with two hidden layers, each consisting of 256 hidden units, and use ELU activation functions for both the token-mixing and channel-mixing MLPs.

The encoder receives a stack of temporal observations at time $t$ over the past $H$ measurements as $\textbf{o}^{\mathrm{p},H}_t=\begin{bmatrix}\textbf{o}^\mathrm{p}_t\ \textbf{o}^\mathrm{p}_{t-1}\;\cdots\;\textbf{o}_{t-H} \end{bmatrix}^T$, allowing the policy to infer context with short-term memory. Specifically, we set $H\!=\!5$, with the policy running at $50~\textrm{Hz}$, providing a memory window of $100~\textrm{ms}$.

The proprioceptive encoder is trained to output a distribution over latent states using variational inference~\cite{higgins2016beta}, supporting exploratory learning and serving as a denoising mechanism to aid domain adaptation. This stochastic latent representation significantly reduces the sim-to-real gap, resulting in smooth and robust real-world control~\cite{nahrendra2023dreamwaq}. 

The latent vector $\textbf{z}^{\mathrm{p}}_t$ is used as input to the multi-modal mixer and for body velocity estimation via an additional estimation layer following the encoder. The body velocity estimation layer is a fully connected layer with 256 hidden units and ELU activation, which predicts the body linear velocity $\hat{\textbf{v}}_t$ using the loss function in Section~\ref{section:estimation_loss}.

\subsection{Multi-Modal Mixer}
The multi-modal mixer network was implemented using an MLP-mixer architecture, where the input features are treated as tokens and the time frames are treated as channels. Both the token-mixing and channel-mixing MLPs consist of two hidden layers with 256 hidden units each and employ ELU activation functions. 

To stabilize learning, we apply layer normalization separately to the proprioceptive and exteroceptive latent features before passing them into the multi-modal mixer. This normalization step ensures that the two modalities are on a comparable scale, thereby facilitating more effective fusion.

The multi-modal mixer was trained end-to-end alongside all other networks in DreamWaQ++, as shown in Fig.~\ref{figure:dreamwaqpp}. However, this setup presents a trade-off in numerical stability during the early stages of training.

A straightforward solution is to balance the weights of the reconstruction loss and KL divergence in $\mathcal{L}_\mathrm{VAE}$. In variational inference, the prior distribution of the latent state is typically assumed to follow a normal distribution. Strong minimization of latent loss during training can help the encoder better approximate the latent space, enhancing training stability. However, this approach risks posterior collapse, wherein the encoder neglects critical input details, impairing the policy’s ability to detect and respond to small environmental obstacles. To address this, we introduced a constrained reparameterization trick, defined as follows:

\begin{equation}
    \textbf{z} \sim N(g_{\mu}(\textbf{x}),g_{\sigma}(\textbf{x})),
\end{equation}
where $\textbf{z}$ is a stochastic latent vector and $\textbf{x}$ is the input to the encoder network $g$. The subscripts $\mu$ and $\sigma$ indicate the outputs of $g(x)$ that correspond to the mean and standard deviation of the latent distribution, respectively. $\textbf{z}$ is sampled from a Gaussian distribution $N(\cdot,\cdot)$ with mean $g_{\mu}(\textbf{x})$ and standard deviation $g_{\sigma}(\textbf{x})$. 

During the reparameterization step, we imposed hard constraints on the standard deviation of the distribution, ensuring $\sigma_\mathrm{min} \leq g_{\sigma}(\textbf{x}) \leq \sigma_\mathrm{max}$. This constraint guarantees numerically stable samples that can be reliably propagated to subsequent network layers. By implementing this simple yet effective solution, we achieve greater training stability without compromising the policy’s final performance. Empirically, we set $\sigma_\mathrm{min}\!=\!0$ and $\sigma_\mathrm{max}\!=\!5$ to facilitate stable training. This constrained reparameterization trick is applied across all encoder networks that use a stochastic layer.

\subsection{Learning Objectives}
We trained the multi-modal context encoder using three losses: an estimation loss, $\mathcal{L}_\textrm{est}$, a proprioceptive variational auto-encoder (VAE) loss, $\mathcal{L}^\textrm{p}_\textrm{VAE}$, and an exteroceptive VAE loss,  $\mathcal{L}^\textrm{e}_\textrm{VAE}$. These losses are combined and included as an auxiliary term in the policy loss.

\subsubsection{Estimation Loss}\label{section:estimation_loss}
The estimation loss is used to train the proprioceptive encoder to explicitly estimate the body velocities of the robot, $\tilde{\textbf{v}}_t$.  The estimation objective was formulated using mean-squared-error (MSE) loss as:
\begin{equation}
  \mathcal{L}_\text{est}=\mathrm{MSE}(\tilde{\textbf{v}}_t,\textbf{v}_t),
\end{equation}
where $\textbf{v}_t$ as the ground-truth (GT) body velocity of the robot in the robot frame. We also adaptively bootstrap $\tilde{\textbf{v}}_t$ during policy training to improve the robustness of the policy~\cite{nahrendra2023dreamwaq}. To avoid exploiting inaccurate estimation in the early stage of training, a bootstrapping probability, $p_\text{boot}\!\in\![0,1]$, is computed by measuring the coefficient of variation, $CV(\cdot)$ of the cumulative rewards $\textbf{R}\in\mathbb{R}^{m\times1}$. The probability is formulated as 
\begin{equation}
  p_\text{boot}= 1 - \tanh(CV(\textbf{R})).
  \label{eqn:adaboot}
\end{equation}

\subsubsection{VAE Loss} 
The multi-modal context encoder is trained using an unsupervised method with two reconstruction tasks. First, the proprioceptive encoder is trained to reconstruct the future observation,  $\tilde{\textbf{o}}_{t+1}$, to encourage the predictive nature of the network. We employ $\beta$-VAE loss for the proprioceptive encoder, formulated as 
\begin{equation}
  \mathcal{L}^\textrm{p}_\text{VAE} = \mathrm{MSE}(\tilde{\textbf{o}}_{t+1},\textbf{o}_{t+1}) + \beta D_\text{KL}(q(\textbf{z}^\textrm{p}_t|\textbf{o}^{\mathrm{p},H}_t)\parallel p(\textbf{z}^\textrm{p}_t)),
\end{equation}
where the first term is the reconstruction loss and the second term is the latent regularization loss expressed with a Kullback-Leibler (KL) divergence operation. The latent regularization is scaled with $\beta=5.0$ to encourage disentanglement~\cite{nahrendra2023dreamwaq,higgins2016beta}. The prior distribution of the proprioceptive context $p(\textbf{z}^\textrm{p}_t)$ is parameterized using a Gaussian distribution and the posterior distribution $q(\textbf{z}^\textrm{p}_t|\textbf{o}^H_t)$ is approximated using a neural network, i.e. via the encoder network. 

Second, the exteroceptive and multi-modal context encoders are trained with an exteroceptive VAE loss, formulated as 
\begin{equation}
  \mathcal{L}^\textrm{e}_\text{VAE} = \mathrm{MSE}(\tilde{\textbf{h}}_{t},\textbf{h}_{t}) + \beta D_\text{KL}(q(\textbf{z}^\textrm{pe}_t|\textbf{o}^\textrm{pe}_t)\parallel p(\textbf{z}^\textrm{pe}_t)),
\end{equation}
where $\textbf{z}^\textrm{pe}_t=f_{\psi_{\mathrm{mix}}}(\textbf{z}^\textrm{p}_t \concat \textbf{z}^\textrm{e}_t)$ is the output of the multi-modal context encoder, as a result of feeding the concatenation of the proprioceptive and exteroceptive context vectors into the mixer network, $f_{\psi_{\mathrm{mix}}}(\cdot)$. $\textbf{o}^\textrm{pe}_t = \textbf{o}^{\textrm{p},H}_t \concat \textbf{o}^{\textrm{e},K}_t$ is the observable proprioception and exteroception. The ground-truth robot-centric height scan, $\textbf{h}_{t}$, is obtained from the simulator, and $\tilde{\textbf{h}}_{t}$ is its reconstruction, which can be obtained via a decoder network that receives $\textbf{z}^\textrm{pe}_t$ as its input.
 
A large value of $\beta$ imposes a strong latent regularization, thus, limiting the reconstruction accuracy, and vice versa. Although it is only a single parameter, tuning $\beta$ is non-trivial and lack of intuition. Therefore, we propose an adaptive $\beta$ scheduling method to ease its tuning procedure by scaling it with a factor, $k$, computed as:
\begin{equation}
    k=\exp{\left(\delta\cdot(\tau-\mathcal{L}_\textrm{recon})\right)},
\end{equation}
where $\delta > 0$ is the learning rate for $k$, $\tau$ is the allowed reconstruction error threshold, and $\mathcal{L}_\textrm{recon}$ is the reconstruction loss. Subsequently, $\beta$ is updated using the following rule:
\begin{equation}
    \beta \leftarrow \begin{cases}
        \beta_\textrm{min} & \text{if } k\beta \leq \beta_\textrm{min}, \\
        k\beta & \text{if } \beta_\textrm{min} \leq k\beta \leq \beta_\textrm{max}, \\
        \beta_\textrm{max} & \text{if } k\beta > \beta_\textrm{max}.
    \end{cases}
\end{equation}
Intuitively, $k$ is updated at every iteration depending on the reconstruction loss of the VAE network. When the reconstruction error exceeds a certain threshold, $\tau$, then $\beta$ is scaled down to allow learning of more accurate reconstruction. In contrast, when the reconstruction error is below the given threshold, $\beta$ is scaled up to allow learning of more disentangled latent representation.

\subsubsection{Contrastive Loss}
Prior works trained an adaptation encoder using a regression loss to explicitly predict environment properties~\cite{kumar2021rma,lee2020learning}. However, this approach might suffer from \textit{realizability gap}~\cite{fu2022deep} caused by insufficient observations to reconstruct the environment properties. To circumvent this issue, we tighten the distribution gap between the learned latent representations of policy's observations and critic's privileged observations, rather than requiring the policy to infer the privileged information via regression. We employ a contrastive learning framework by matching the distribution of the privileged latent features used for the critic with the latent features inferred from partial observations used for the actor within an asymmetric actor-critic setup. We define the contrastive loss as: 
\begin{align}
    \mathcal{L}_\mathrm{contrastive} = &\lambda\left\|\textbf{z}^\textrm{pe}_t - g_{\theta_\mathrm{h}}(\textbf{h}_t)\right\|_2^2 \nonumber \\
    &+ (1-\lambda)\left\|\max(0, m-(\textbf{z}^\textrm{pe}_t - \textbf{z}^\textrm{random}_t))\right\|_2^2,\label{eq:contrastive_loss}
\end{align}
where $g_{\theta_\mathrm{h}}(\textbf{h}_t)$ is the encoded ground-truth height scan, which is used as the positive anchor for the contrastive loss. Meanwhile, $\textbf{z}^\textrm{random}_t$ is a random latent feature sampled from $\mathcal{U}[-1.0,1.0]$, which is used as the negative anchor. The parameters $m\!\in\!\mathbb{R}^+$ and $\lambda\!\in\![0, 1]$ are the margin for the negative pair separation and scaling factor, respectively. This contrastive loss forces the multi-modal latent feature to match the encoded ground-truth height scan, while also distancing the latent feature from an unstructured representation labeled by the uniformly random latent feature.


\subsection{Skill Discovery}\label{section:skill_discovery}

We incorporate an unsupervised RL objective through mutual information (MI) maximization for promoting skill discovery. This objective allows the emergence of novel behaviors while preserving stable behaviors induced by the handcrafted reward functions. Specifically, we maximize the MI between visited states and the latent variable inferred by the multi-modal context encoder. 

The MI objective is introduced as a regularization term in the PPO loss function. We call this objective as \textit{versatility gain}, which seeks to be maximized for inducing versatile locomotion behaviors. Thus, the versatility gain can balance exploration, exploitation, and reconstruction. The versatility gain is defined as:
\begin{equation}\label{eq:versatility_gain}
    \mathcal{G}_\textrm{versatility}=\mathcal{I}(\textbf{o}^\textrm{pe}_t;\textbf{z}^\textrm{pe}_t)=\mathcal{H}(\textbf{z}^\textrm{pe}_t)-\mathcal{H}(\textbf{z}^\textrm{pe}_t|\textbf{o}^\textrm{pe}_t),
\end{equation}
where $\mathcal{I}(\cdot;\cdot)$, $\mathcal{H}(\cdot)$, and $\mathcal{H}(\cdot|\cdot)$ are the mutual information, Shannon entropy, and conditional entropy operators, respectively. \eqref{eq:versatility_gain} comprises two terms that were essential for training. The first term maximizes the variation of the inferred latent variables, thus, promoting the variation of skills that can be obtained during policy learning. The second term minimizes the entropy of the latent states given an observation, thus, acting as a denoising operation to filter out noisy observations by clustering intrinsically similar observations into a similar latent representation.

Generally, the encoder is trained to minimize the KL divergence between $\textbf{z}^\textrm{pe}_t$ and $\textbf{o}^\textrm{pe}_t$, i.e. $\mathcal{L}_\textrm{encoder} \approx D_\textrm{KL}\left(\textbf{z}^\textrm{pe}_t | \textbf{o}^\textrm{pe}_t\right)$, effectively compressing raw observations while maintaining the original data distribution. Subsequently, jointly training the networks using $\mathcal{G}_\textrm{versatility}$ and $\mathcal{L}_\textrm{encoder}$ maximizes:

\begin{align}
    \mathcal{J} &\triangleq \mathcal{G}_\textrm{versatility} - \lambda_\mathrm{e} \mathcal{L}_\textrm{encoder} \nonumber \\ 
                &= \mathcal{I}(\textbf{o}^\textrm{pe}_t;\textbf{z}^\textrm{pe}_t) - \lambda_\textrm{e} D_\textrm{KL}\left(\textbf{z}^\textrm{pe}_t | \textbf{o}^\textrm{pe}_t\right) \nonumber \\
                &= \mathcal{H}(\textbf{z}^\textrm{pe}_t)-\mathcal{H}(\textbf{z}^\textrm{pe}_t|\textbf{o}^\textrm{pe}_t) + \lambda_\textrm{e} \left[ \mathcal{H}(\textbf{z}^\textrm{pe}_t,\textbf{o}^\textrm{pe}_t) - \mathcal{H}(\textbf{z}^\textrm{pe}_t)\right] \nonumber \\
                &= \mathcal{H}(\textbf{z}^\textrm{pe}_t)-\mathcal{H}(\textbf{z}^\textrm{pe}_t|\textbf{o}^\textrm{pe}_t) + \lambda_\textrm{e} \left[\mathcal{H}(\textbf{z}^\textrm{pe}_t|\textbf{o}^\textrm{pe}_t) + \mathcal{H}(\textbf{o}^\textrm{pe}_t) - \mathcal{H}(\textbf{z}^\textrm{pe}_t)\right] \nonumber \\
                &= \left(1 - \lambda_\textrm{e}\right) \mathcal{H}(\textbf{z}^\textrm{pe}_t) - \left(1 - \lambda_\textrm{e}\right) \mathcal{H}(\textbf{z}^\textrm{pe}_t|\textbf{o}^\textrm{pe}_t) + \lambda_\textrm{e} \mathcal{H}(\textbf{o}^\textrm{pe}_t), \label{eq:combined_objective}
\end{align} 
where $\mathcal{H}(\cdot,\cdot)$ is a cross-entropy operation and $\lambda_\textrm{e}\in\mathbb{R}^+$ is the scaling factor for $\mathcal{L}_\textrm{encoder}$. \eqref{eq:combined_objective} shows that choosing $\lambda_\textrm{e}=1$ leads to entropy maximization on the state visitation that subsequently promotes policy exploration and skill discovery during training. Furthermore, choosing $\lambda_\textrm{e}<1$ maximizes~$\mathcal{H}(\textbf{z}^\textrm{pe}_t)$ and minimizes~$\mathcal{H}(\textbf{z}^\textrm{pe}_t|\textbf{o}^\textrm{pe}_t)$, effectively diversifying the distribution of $\textbf{z}^\textrm{pe}_t$ while compressing $\textbf{o}^\textrm{pe}_t$. In practice, we set $\lambda_\textrm{e}=0.1$ for our experiments.
\section{Implementation Details}
\subsection{Simulation}
We utilized NVIDIA Isaac Gym Preview 3~\cite{makoviychuk2021isaac} as the simulator to train the controller and multi-modal context encoder networks, with training environments based on the Legged Gym library~\cite{rudin2021learning}. Domain randomization was applied across $3,\!500$ agents, completing training in approximately 11 hours on an NVIDIA A5000 GPU. The trained multi-modal context encoder and policy networks were then deployed without any fine-tuning on a physical robot or in the Gazebo simulator for the evaluations presented in this paper.

\subsection{Low-Level Control}
The policy and multi-modal context encoder networks run synchronously while processing asynchronous observations. Proprioceptive measurements are sampled at $200~\textrm{Hz}$, and exteroceptive measurements at $10~\textrm{Hz}$, with the controller integrating the latest measurements at $50~\textrm{Hz}$. To enhance robustness against asynchronous observations, latency randomization was applied during training. Detailed information on randomized latency for all measurements is available in Table~\ref{table:domain_randomization}.

The policy network generates target joint positions at $50~\textrm{Hz}$, which are then sent to a low-level proportional-derivative (PD) controller operating at $200~\textrm{Hz}$. Within the PD controller, target joint positions are converted into torque commands using proportional ($K_p$) and derivative ($K_d$) gains of $25$ and $0.7$, respectively. We made a custom interface using Pybind to access the Unitree SDK via our Python script that runs the RL policy and to send the target joint positions to the Unitree SDK using a ROS system. The Pybind interface then converts the target joint positions into torque commands, which are subsequently transmitted to the low-level motor controller.
\subsection{Domain Randomization}
    We randomized multiple physical properties of the robot and environment to facilitate sim-to-real transfer. Additionally, we employed Roll-Drop~\cite{campanaro2023roll} to encourage exploration and enhance robustness alongside physics randomization. Table~\ref{table:domain_randomization} summarizes the details of the physical properties and their randomization ranges.

    Additionally, we applied domain randomization to account for potential data asynchrony between proprioceptive and exteroceptive sensors caused by system delays. Such delays may arise from factors including data transmission latency, multithreaded execution on the robot, and hardware-induced timing jitter. These issues can violate the Markov property assumed in standard policy learning frameworks.

    To mitigate this, we randomly delay proprioceptive observations within a range of $[0, 15]~\mathrm{ms}$ during training. This strategy encourages the policy to treat minor mismatches between proprioceptive and exteroceptive inputs as observation noise, thereby increasing its robustness to sensor delays during deployment.

    While it is theoretically possible to extend the randomization range to further improve robustness, doing so can lead to overly conservative policies and degrade performance. Therefore, we set the maximum delay to $15~\mathrm{ms}$, which we found to be empirically appropriate given the computational constraints of our onboard hardware.

    \begin{table}[t!]
    \centering
    \caption{Domain randomization ranges applied in the simulation.}
    \label{table:domain_randomization}
    \begin{center}
    \begin{tabular}{lccc}
    \hline\hline
    Parameter                        & Randomization range & Unit \\\hline
    Payload                     & $[-1, 2]$ & $\mathrm{~kg}$     \\ 
    $K_p$ factor   & $[0.9, 1.1]$ & $\mathrm{~Nm~rad^{-1}}$    \\ 
    $K_d$ factor        & $[0.9, 1.1]$ & $\mathrm{~Nms~rad^{-1}}$   \\ 
    Motor strength factor        & $[0.9, 1.1]$ & $\mathrm{~Nm}$   \\ 
    Center of mass shift        & $[-50,50]$ & $\mathrm{~mm}$    \\
    Friction coefficient        & $[0.2, 1.25]$ & -    \\
    System delay                        & $[0.0, 15.0]$ &  $\mathrm{~ms}$   \\

    \hline\hline
    \end{tabular}
    \end{center}
    \vspace{-0.cm}
    \end{table}

\subsection{Adversarial Observations}\label{section:adversarial}
    During training, we injected noise into the proprioceptive and exteroceptive observations to make the policy be robust against noisy real-world observations. For the proprioceptive observations, a uniform noise was injected at each time step. For the exteroceptive data, we defined three different noise ranges, constituting low, medium, and high noise levels. The proportion of robots that operate with these exteroceptive noise scales during training was set to $30\%$, $50\%$, and $20\%$, for the low, medium, and high noise levels, respectively.

    To handle erroneous extrinsic calibration between the exteroceptive sensor and the robot body frame, we also applied sensor alignment bias at the beginning of each episode to simulate extrinsic calibration error. This error encompasses biases in both the position and orientation of the sensor frame with respect to the robot body frame. The biases were sampled from a uniform distribution and consistently applied to the exteroceptive measurements throughout the episode. The ranges of the observation noises and sensor alignment biases are summarized in Table~\ref{table:observation_noise}.

    \begin{table}[t!]
    \centering
    \caption{Noise and sensor alignment bias parameters injected into the observation for the policy network during training.}
    \label{table:observation_noise}
    \begin{center}
    \begin{tabular}{lccc}
    \hline\hline
    Observation                        & Noise range ($\mu$) & Unit \\\hline
    Joint position  &                      $[-0.01, 0.01]$ & $\mathrm{rad}$     \\ 
    Joint velocity  &                      $[-1.5, 1.5]$ & $\mathrm{rad/s}$    \\ 
    Body linear velocity &                 $[-0.1, 0.1]$ & $\mathrm{m/s}$   \\ 
    Body angular velocity &                $[-0.2, 0.2]$ & $\mathrm{rad/s}$   \\ 
    Gravity vector        &                $[-0.05, 0.05]$ & $\mathrm{m/s^{2}}$    \\ 
    Exteroceptive measurement (low)  &     $[0.0, 0.03]$ & $\mathrm{m}$\\
    Exteroceptive measurement (medium)  &     $[0.03, 0.1]$ & $\mathrm{m}$    \\
    Exteroceptive measurement (high)  &     $[0.1, 0.3]$ & $\mathrm{m}$    \\
    Exteroceptive bias roll             & $[-0.2, 0.2]$ &  $\mathrm{rad}$   \\
    Exteroceptive bias pitch           & $[-0.15, 0.15]$ &  $\mathrm{rad}$   \\
    Exteroceptive bias yaw              & $[-0.1, 0.1]$ &  $\mathrm{rad}$   \\
    Exteroceptive bias $\mathrm{x}$             & $[-0.1, 0.1]$ &  $\mathrm{m}$   \\
    Exteroceptive bias $\mathrm{y}$             & $[-0.1, 0.1]$ &  $\mathrm{m}$   \\
    Exteroceptive bias $\mathrm{z}$             & $[-0.1, 0.1]$ &  $\mathrm{m}$   \\

    \hline\hline
    \end{tabular}
    \end{center}
    \vspace{-0.cm}
    \end{table}

\subsection{Privileged States}
    We utilized a robot-centric local height map as the privileged exteroception, sampled around the robot in a 2.5D grid where each cell represents terrain height. The grid dimensions are $w\!=\!1.1~\mathrm{m}$ by $h\!=\!1.7~\mathrm{m}$, with the first row positioned $0.9~\mathrm{m}$ ahead of the robot’s body frame, similar to the 3D voxel grid used for the multi-modal encoder input. The grid resolution is set to $5~\mathrm{cm}$.

    For privileged proprioception, we utilized ground-truth data and simulation measurements, including body-centric linear and angular velocities, gravity vector, joint positions and angular velocities, external disturbance forces and torques, foot positions relative to the robot's body frame, and physical properties of the robot. The physical properties include friction, motor damping, motor stiffness, motor strength ratio, additional payload, and the robot's center of mass shift.

\subsection{Reward Functions}\label{training:reward_functions}
    We used the exact same rewards in~\cite{nahrendra2023dreamwaq}. On top of these rewards, we employed a reward curriculum to exponentially anneal some style rewards, i.e.~joint torque, joint velocity, joint acceleration, action rate, and smoothness rewards. We observed that although these style rewards are important to ensure sim-to-real transfer, they can lead to suboptimal policies that make the robot only learn basic locomotion skills. This is because the agent may become overly focused on maximizing immediate rewards induced by these joint-level regularization rewards. Therefore, we gradually annealed the weight of the regularization rewards to allow the agent to explore a wider range of behaviors and styles after it has learned the basic locomotion task. The annealing follows the following rule:

    \begin{equation}
        w_{i+1}=\lambda w_{i},
    \end{equation}
    where $w$ is the reward weight, $i$ is the learning iteration, and $\lambda$ is the annealing rate. We set $\lambda\!=0.998$
    and $w_0$ for the selected skill rewards are summarized in Table~\ref{table:reward_init}.

    \begin{table}[t!]
    \centering
    \captionsetup{singlelinecheck=false}
    \caption{Initial weight $w_0$ for selected style rewards that were annealed using the reward curriculum.}
    \label{table:reward_init}

    \begin{center}
    \begin{tabular}{lll}
    \hline \hline
    Reward           & Weight ($w_0$)\\ \hline 
    Joint torque & $-5\!\times\!10^{-6}$ \\
    Joint velocity & $-6\!\times\!10^{-6}$ \\
    Joint acceleration & $-7.5\!\times\!10^{-8}$ \\
    Action rate & $-1.5\!\times\!10^{-5}$ \\
    Smoothness & $-1.5\!\times\!10^{-5}$ \\
    \hline \hline
    \end{tabular}
    \end{center}
    \end{table}

\section{Experimental Results}
    \subsection{Hardware Settings}\label{section:hardware}
\begin{figure}[t!]
	\centering 
	\begin{subfigure}[b]{0.48\textwidth}
		\includegraphics[width=1.0\textwidth]{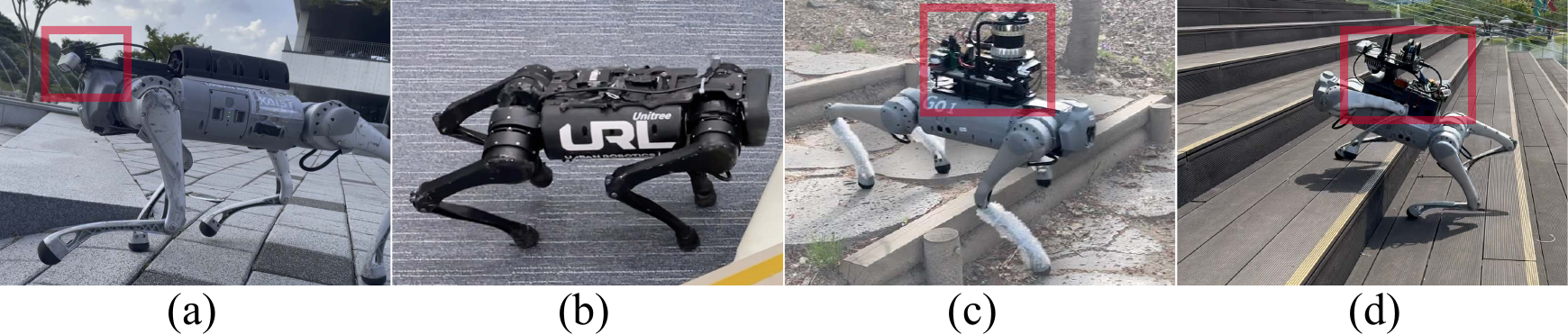}
	\end{subfigure}
	\caption{Hardware setup for robots (a)~R1, (b)~R2, (c)~R3, and (d)~R4. R1 was equipped with an Intel RealSense D435f camera, R2 was not equipped with any exteroceptive sensor and used only for blind locomotion, R3 was equipped with an Ouster OS-01 LiDAR, and R4 was equipped with two Livox Mid-360 LiDARs.}
	\label{figure:hardware}
\end{figure}

All networks were trained to control a Unitree Go1~\cite{unitreego1} robot. For the experiments, we used robots with different exteroception configurations, as shown in Fig.~\ref{figure:hardware}. Robots R1, R3, and R4 are Unitree Go1 robots with different exteroception setups, while Robot R2 is a Unitree A1 robot without exteroception.

Robot R1 was equipped with an Intel RealSense D435f camera tilted $45^\circ$ downward, streaming data at $15~\textrm{Hz}$ to the onboard Jetson Xavier NX. To protect cables from potential damage during falls, a canopy was added, increasing the payload by approximately $0.5~\textrm{kg}$.

Robot R2, a Unitree A1 without exteroception, was used to compete against Robot R1. The blind locomotion policy, DreamWaQ~\cite{nahrendra2023dreamwaq} was deployed on the A1 robot because it has similar morphology and motor properties to the Go1, while offering higher torque. Consequently, the Unitree A1 is expected to perform comparably, if not better, than the Unitree Go1 with DreamWaQ, as it can apply additional torque to manage collisions during stair climbing.

\begin{figure}[t!]
        \centering 
        \begin{subfigure}[b]{0.48\textwidth}
            \includegraphics[width=1.0\textwidth]{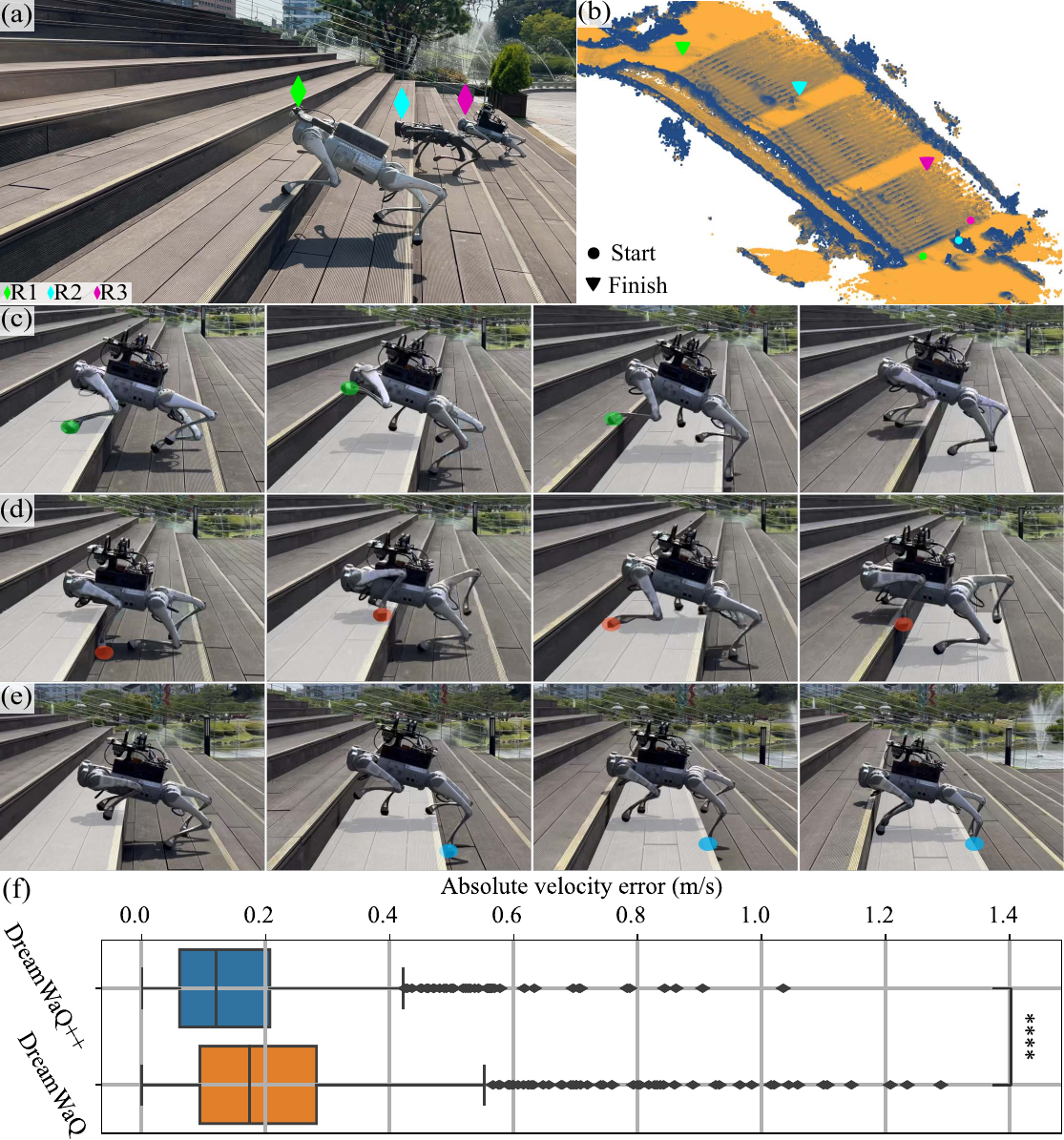}
        \end{subfigure}
            \vspace{-0.2cm}
        \caption{(a)~A head-to-head race between the proposed controller against baselines. (b)~3D map visualization of the race environment. Asynchronous stair-climbing experiments were conducted to ensure fairness, where the robot was controlled using (c)~DreamWaQ++, (d)~DreamWaQ, and (e)~Unitree Go1's built-in controller. The white mask on the stair indicates the same stair plate that each robot interacted with over the four successive snapshots. (f)~Velocity estimation error comparison in the stair-climbing task. The **** annotation indicates the significance level computed using a paired $t$-test.}
        \label{figure:results_stairs}
    \end{figure}

Robot R3 was equipped with an Intel NUC PC and an Ouster OS-01 LiDAR. This configuration was used to evaluate the generalization and transferability of the learned controller on a robot with different exteroceptive sensors, resulting in an additional payload of approximately $3.0~\textrm{kg}$. Finally, a fourth robot, R4, equipped with an Intel NUC PC and two Livox Mid-360 LiDARs, was utilized for further experiments in an asynchronous race setting shown in Figs.~\ref{figure:results_stairs}(c)-(e). The additional payloads on R3 and R4 are approximately $2.5~\textrm{kg}$.

    \subsection{Fast Locomotion over Obstacles}

    \subsubsection[Head-to-head Racing Across Stairs]{Head-to-head Racing Across Stairs\protect\mvtwofoot}
    We benchmarked the proposed controller against DreamWaQ~\cite{nahrendra2023dreamwaq} (a baseline blind locomotion controller) and the robot's built-in perceptive controller~\cite{unitreego1} in a head-to-head stair-climbing race as shown in Fig.~\ref{figure:results_stairs}(a). 

    The experimental setup consisted of fifty stairs, and we used robots R1, R2, and R3 as described in Section~\ref{section:hardware}. Robots R1, R2, and R3 were controlled by DreamWaQ++, DreamWaQ, and Unitree's built-in controller, respectively. All robots started from the same point before the stairs, except for R3, which was placed one stair ahead due to the built-in controller's difficulty with the height of the first step. For safety reasons, all robots were manually controlled by a human operator.

    Robot R1 quickly outperformed the others, gaining a lead shortly after the starting point, as shown in Fig.~\ref{figure:results_stairs}(b). To mitigate frequent stumbles, robot R2 was commanded at a linear velocity of $1.2~\mathrm{m/s}$ but continued to experience tracking errors due to frequent missteps on stair edges. In contrast, robot R1 adaptively adjusted its gait and foot placement, enabling faster traversal at a lower velocity command of approximately $1.0~\mathrm{m/s}$. Robot~R3, however, struggled with stair climbing due to latency in its perceptive controller’s reaction time, which depends on a local environmental map and limits both speed and agility.

    The race concluded when one robot reached the final stair to assess overall reachability (Fig.~\ref{figure:results_stairs}(b)). Robot R1 completed the race in $35~\mathrm{s}$, covering a horizontal distance of approximately $30.03~\mathrm{m}$ and a height of $7.38~\mathrm{m}$. By the same timestamp, R2 had covered about $20.05~\mathrm{m}$ horizontally and $5.44~\mathrm{m}$ in height, while R3 failed to finish after stumbling and covering only $6.38~\mathrm{m}$ horizontally and $2.44~\mathrm{m}$ in height. This experiment highlights the superior agility of the proposed controller in handling continuous obstacles.

    \subsubsection[Asynchronous Race]{Asynchronous Race\protect\mvtenfoot}
    We conducted additional asynchronous experiments using robot R4 (Fig.~\ref{figure:hardware}(d)) with autonomous navigation to ensure consistent path planning during stair climbing. This experiment aimed to validate the performance of DreamWaQ++ in a more controlled setting, where the same robot hardware was tested using both DreamWaQ and DreamWaQ++. Key snapshots highlighting the robot's movements are shown in Figs.~\ref{figure:results_stairs}(c)-(e).

    A discernible behavior between DreamWaQ++ and DreamWaQ is in their gait adaptation. DreamWaQ++ raises the body and extends foot swing to step securely on the stairs (Fig.~\ref{figure:results_stairs}(c)), whereas DreamWaQ often collides and drags the foot along the stair edge before stepping (Fig.~\ref{figure:results_stairs}(d)), resulting in reduced efficiency and more frequent stumbles. The built-in controller uses a fixed gait, lacking spatial memory and adaptability, leading to rear leg stumbles (Fig.~\ref{figure:results_stairs}(e)).

    We also measured state estimation errors for DreamWaQ++ and DreamWaQ during these experiments. This error, defined as the absolute difference between ground truth and estimated velocities, was computed using a LiDAR odometry algorithm~\cite{xu2022fast}. As shown in Fig.~\ref{figure:results_stairs}(f), DreamWaQ++ demonstrated significantly lower estimation errors, indicating superior position estimation and gait adaptation—crucial for efficient stair climbing and enhanced stability and robustness.

    \subsection[Obstacle Awareness]{Obstacle Awareness\protect\mvthreefoot}

    \begin{figure}[t!]
        \centering 
        \begin{subfigure}[b]{0.48\textwidth}
            \includegraphics[width=1.0\textwidth]{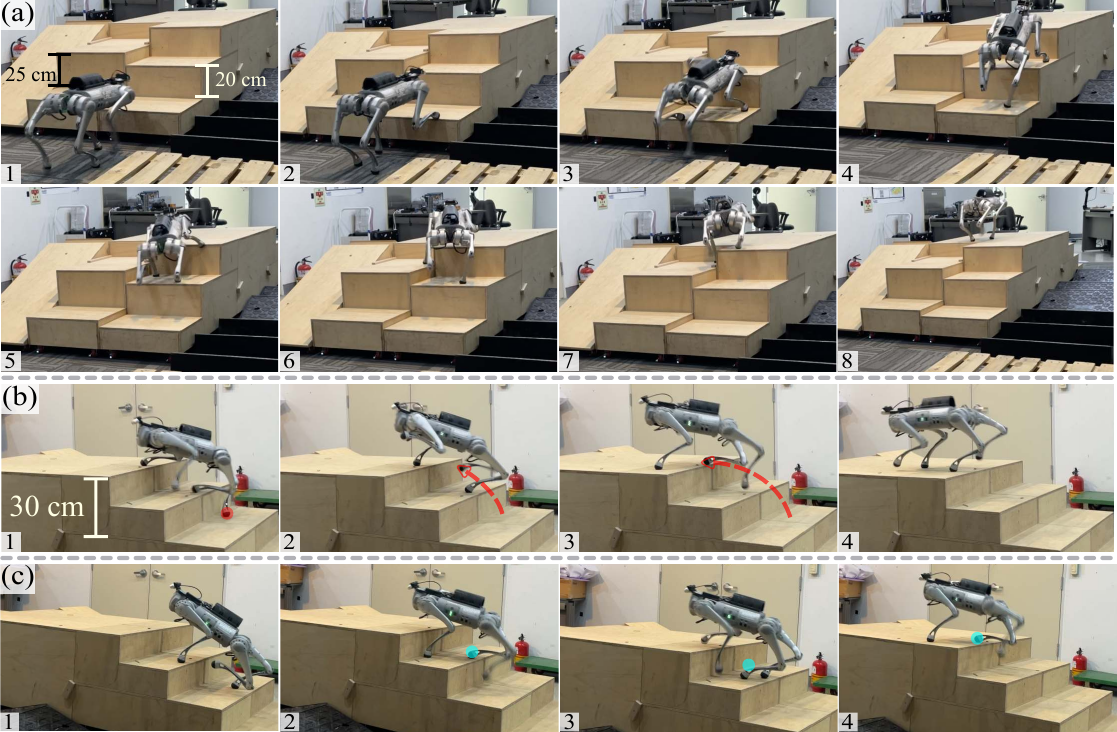}
        \end{subfigure}
        \captionsetup{font=footnotesize}
        \caption{(a)~Affordance-aware locomotion when ascending stairs with rise of $25~\textrm{cm}$ on the left and $20~\textrm{cm}$ on the right side of the robot. (b)~Emergent behavior to quickly and efficiently climb stairs with long foot swing motion, compared with a regular case (c)~where the robot could not overcome two stair steps at once because the rear foot was located around the middle of the stair step.}
        \label{figure:results_awareness}
    \end{figure}

    \subsubsection{Obstacle Negotiation}
        In Fig.~\ref{figure:results_awareness}(a), the controller demonstrated affordance-awareness when navigating terrains of varying difficulty. The robot was given a forward velocity command of $0.6~\mathrm{m/s}$ with a zero yaw rate command. Initially, when commanded to move toward the middle of the stairs (Fig.~\ref{figure:results_awareness}(a-2)), the robot moved toward the lower stair rise on the right side, opting for a less risky path. Subsequently, as the left and right stairs overlapped, a lower step appeared on the left side (Fig.~\ref{figure:results_awareness}(a-6)). The robot quickly adapted its path toward these easier steps, resisting the zero yaw rate command, thereby demonstrating the controller’s ability to learn and perceive the affordances of different obstacles.

    \subsubsection{Foot Swing Adaptation}
        Fig.~\ref{figure:results_awareness}(b) illustrates the performance of the proposed controller on stairs with a $15~\textrm{cm}$ rise. Typically, the controller directs the robot to swing its feet, stepping on each stair sequentially. However, when a foot approaches the edge of a stair, the robot extends its swing phase (indicated by red arrows in Fig.~\ref{figure:results_awareness}(b)), allowing the rear foot to overcome a combined rise of $30~\textrm{cm}$. In contrast, as shown in Fig.~\ref{figure:results_awareness}(c), the robot steps normally when the rear foot is positioned near the middle of the initial stair step. This adaptive behavior suggests that the controller retains a form of memory of the structure beneath the robot, leveraging fused information from the network architecture.

    \subsubsection{Quantitative Performance Assessment}
        \begin{figure}[t!]
            \centering 
            \begin{subfigure}[b]{0.48\textwidth}
                \includegraphics[width=1.0\textwidth]{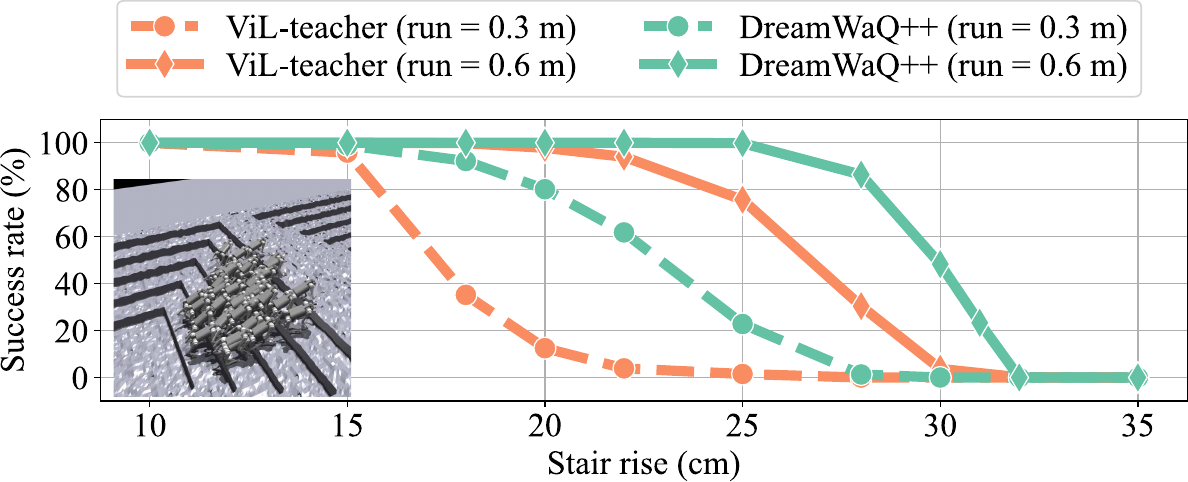}
            \end{subfigure}
            \captionsetup{font=footnotesize}
            \caption{Success rate~(SR) comparison measured on each algorithm by simulating $1,\!000$ robots to climb sequence of stairs. The SR is defined as the percentage of the number of robots that reached the last stair within $10~\mathrm{s}$ over the total number of robots.}
            \label{figure:results_stairs_quant}
        \end{figure}

        We compared DreamWaQ++ with a visual locomotion controller based on ViNL~\cite{kareer2023vinl}, training the baseline with the same parameters as DreamWaQ++ but without the navigation pipeline from~\cite{kareer2023vinl}. Additionally, we used only the teacher network to achieve upper-bound performance, noting that this network has access to the ground truth height map around the robot, which we refer to as \textit{ViL-teacher}.

        We simulated $1,000$ robots climbing stairs with increasing rises, as shown in Fig.~\ref{figure:results_stairs_quant}. Using a stair run of $0.3~\mathrm{m}$—a common real-world dimension—and an extended run of $0.6~\mathrm{m}$ to represent low-slope, high-rise obstacles, the results in Fig.~\ref{figure:results_stairs_quant} indicate that DreamWaQ++ achieved success rates $20\!-\!40\%$ higher than the baseline with ground truth height map access. This finding suggests that accurate height map information alone is insufficient for high-performance locomotion, similar to animals' ability to climb without constant visual feedback.

        This improvement is attributed to versatile skill learning promoted by the proposed versatility gain function~$\mathcal{L}_\mathrm{versatility}$, which serves as an intrinsic reward encouraging exploration. Without this gain, the policy lacks exploratory behavior, often resulting in a conservative approach that fails to handle unseen terrains effectively.
    \subsection[Uncertainty Awareness and Adaptation]{Uncertainty Awareness and Adaptation\protect\mvfourfoot}
    \begin{figure}[t!]
        \centering 
        \begin{subfigure}[b]{0.48\textwidth}
            \includegraphics[width=1.0\textwidth]{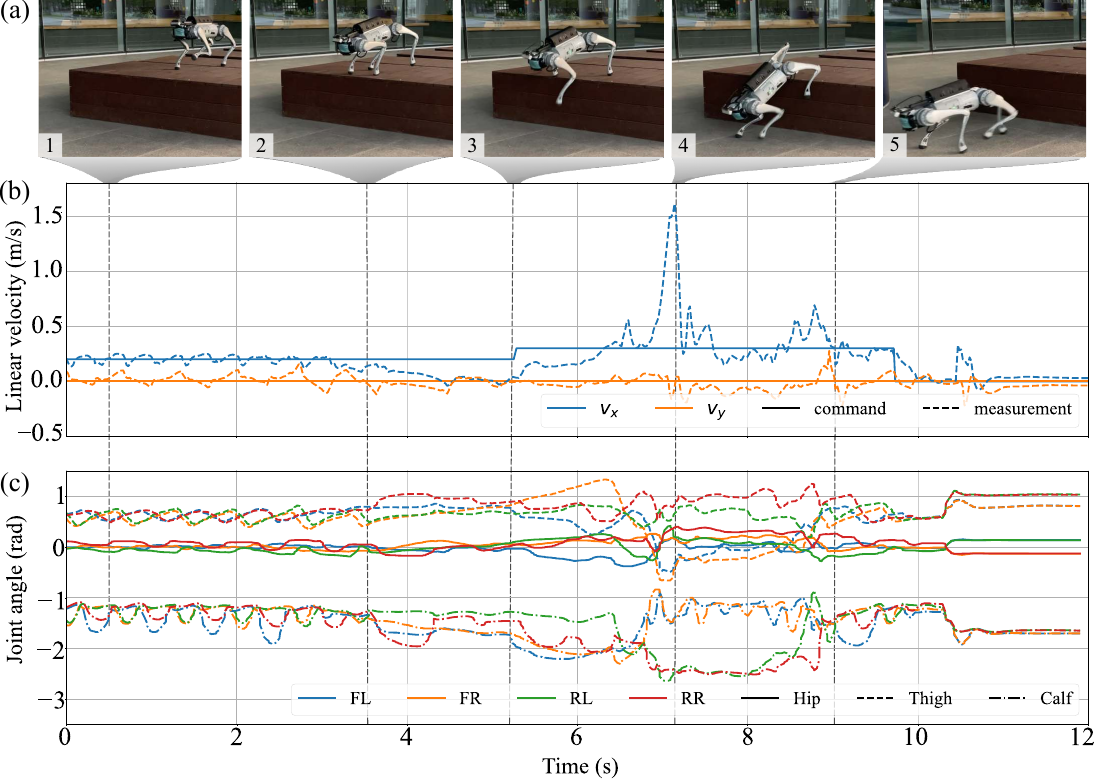}
        \end{subfigure}
        \caption{An emergent probing skill enables the robot to check the upcoming terrain when it poses a high risk and uncertainty. (a)~A sequence of the robot's movement to probe the upcoming terrain. (b)~Corresponding velocity commands and estimation, showing how the controller resists the given command and allocates time for the robot to check for the terrain. (c)~Significant knee flexion-extension~(KFE) motions indicated by a sudden change in the calf joint angle, revealing the emergent adaptive behavior as a novel probing skill.}
        \label{figure:results_probing}
    \end{figure}

    Fig.~\ref{figure:results_probing} demonstrates an emergent locomotion behavior of the proposed controller when traversing terrains with significant height differences. Figs.~\ref{figure:results_probing}(a)-(c) visualize snapshots of the robot's motion, commanded and estimated base velocity, and joint angles, respectively. When confronted with a stage featuring a large elevation difference, the controller could not accurately gauge the terrain height near the robot's front feet. In response, the robot deliberately stops before the ridge (Fig.~\ref{figure:results_probing}(a-2)) and orchestrates its feet to probe the terrain characteristics (Fig.~\ref{figure:results_probing}(a-3)). Upon detecting solid ground, the robot continues moving forward and confidently descends from the edge of the stage (Fig.~\ref{figure:results_probing}(a-4)). Subsequently, the robot spreads its rear legs and uses them as anchors, reducing the impact on the front legs upon landing.

    \subsection{Out-of-distribution Adaptation}
\subsubsection[Reacting to Sudden Changes of Foothold]{Reacting to Sudden Changes of Foothold\protect\mvfivefoot}
\begin{figure}[t!]
    \centering 
    \begin{subfigure}[b]{0.48\textwidth}
        \includegraphics[width=1.0\textwidth]{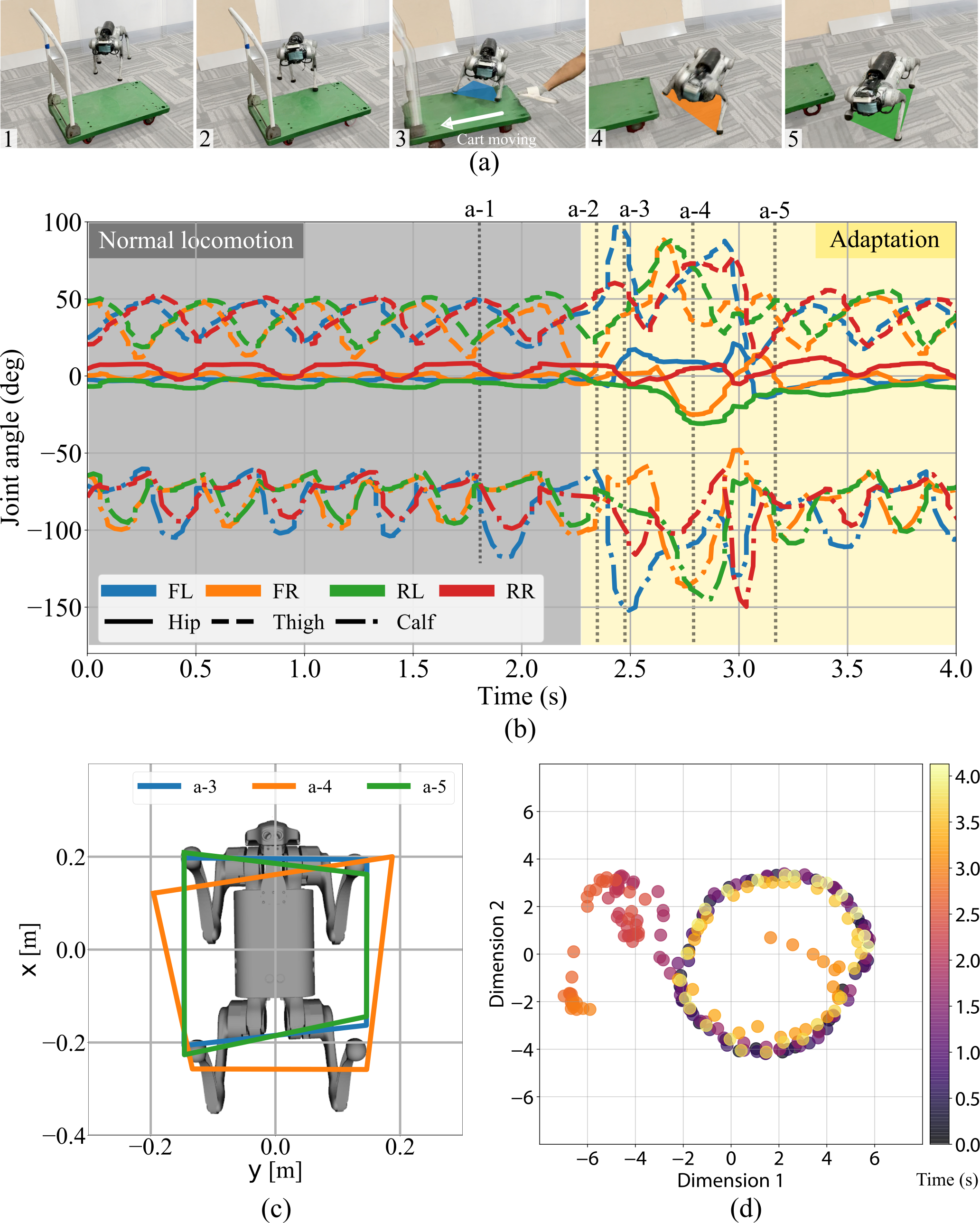}
    \end{subfigure}
    \caption{Adaptation to sudden changes of foothold. (a)~The robot is externally disturbed by quickly removing the platform it is stepping on. (b)~An abrupt change in the robot's perception made the controller rapidly alter the robot's joints at around $t\!=\!2.5~\mathrm{s}$ to (c)~enlarge the robot's support polygon for ensuring a safe and stable landing. (d)~A 2D embedding visualization using \textit{pairwise controlled manifold approximation projection} (PaCMAP)~\cite{wang2021understanding} shows how the multi-modal context dynamically changes over time and capture changes in the environment, providing informative contexts to swiftly adapt the policy.}
    \label{figure:results_adaptation}
\end{figure}

We evaluated the controller's adaptability to unexpected environmental changes, such as deformable and movable surfaces, which were not encountered during training. In Fig.~\ref{figure:results_adaptation}(a), the robot initially moves toward a movable cart. Upon stepping onto the cart, an abrupt kick is applied, propelling the cart away from the robot.

The controller’s swift response, shown in Fig.~\ref{figure:results_adaptation}(a-4), involved manipulating the front hip joints (Fig.~\ref{figure:results_adaptation}(b-4)) to create a support polygon approximately $20.12\%$ larger than in normal locomotion (Fig.~\ref{figure:results_adaptation}(c)), ensuring a safe landing after the platform was unexpectedly removed.

Fig.~\ref{figure:results_adaptation}(d) visualizes the multi-modal contexts, forming a circular pattern corresponding to foot motion events in Fig.~\ref{figure:results_adaptation}(a). When the cart is kicked at $t\!=\!2.4~\mathrm{s}$ (Fig.~\ref{figure:results_adaptation}(a-3)), the contexts form a new cluster (upper left of Fig.~\ref{figure:results_adaptation}(d)). Subsequently, a distinct cluster appears (bottom left of Fig.~\ref{figure:results_adaptation}(d)) at Fig.~\ref{figure:results_adaptation}(a-4), with fewer embeddings as the policy quickly handles the situation. The embeddings then return to the original circular pattern after the robot lands safely, resuming normal locomotion at $t\!=\!3.2~\mathrm{s}$.

\subsubsection[Climbing Over a Steep Slope]{Climbing Over a Steep Slope\mvsixfoot}
\begin{figure}[t!]
    \centering 
    \begin{subfigure}[b]{0.48\textwidth}
        \includegraphics[width=1.0\textwidth]{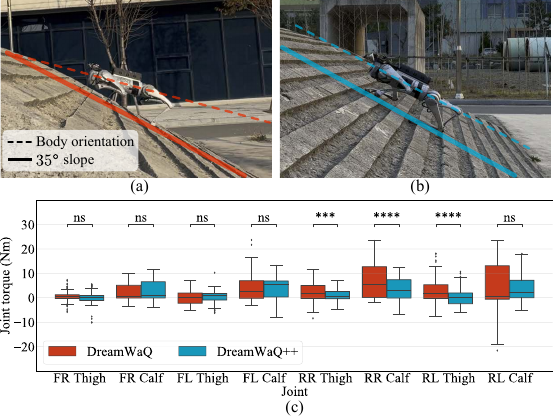}
    \end{subfigure}
    \caption{Comparison of torque exertions when climbing a $35^\circ$ slope using (a)~DreamWaQ and (b)~DreamWaQ++. The annotations on top of the boxplot in~(c) indicate the significance level measured using a paired $t$-test method}
    \label{figure:results_slope}
\end{figure}

Fig.~\ref{figure:results_slope}(c) compares the torque exertion of controllers trained with DreamWaQ and DreamWaQ++ (see Figs.~\ref{figure:results_slope}(a) and~(b)). Both controllers were trained exclusively on rough slopes up to $10^\circ$, while in this experiment, the robot was tasked with climbing a $35^\circ$ slope. DreamWaQ's policy attempts to maintain a flat body orientation (Fig.~\ref{figure:results_slope}(b)), as it was trained with a blind locomotion controller to generalize across various terrains. Although broadly effective, this approach results in conservative behavior and greater torque exertion on the rear legs (Fig.~\ref{figure:results_slope}(a)) to uphold the flat base pose.

In contrast, the DreamWaQ++ policy employs a crawling gait with a reduced body height relative to the slope surface, aligning the robot’s base orientation with the slope inclination, which enhances stability and reduces torque on the rear legs, as shown in Fig.~\ref{figure:results_slope}(b). DreamWaQ++'s terrain perception enables flexible gait adaptation rather than a conservative approach. Notably, rear leg torque in DreamWaQ++ is approximately $1.5$ times lower than in DreamWaQ, demonstrating DreamWaQ++'s superior out-of-distribution adaptation.

    \subsection[Scalability to Other Platforms]{Scalability to Other Platforms\protect\mvsevenfoot}\label{supp:scalability}
\begin{figure}[t!]
	\centering 
	\begin{subfigure}[b]{0.48\textwidth}
		\includegraphics[width=1.0\textwidth]{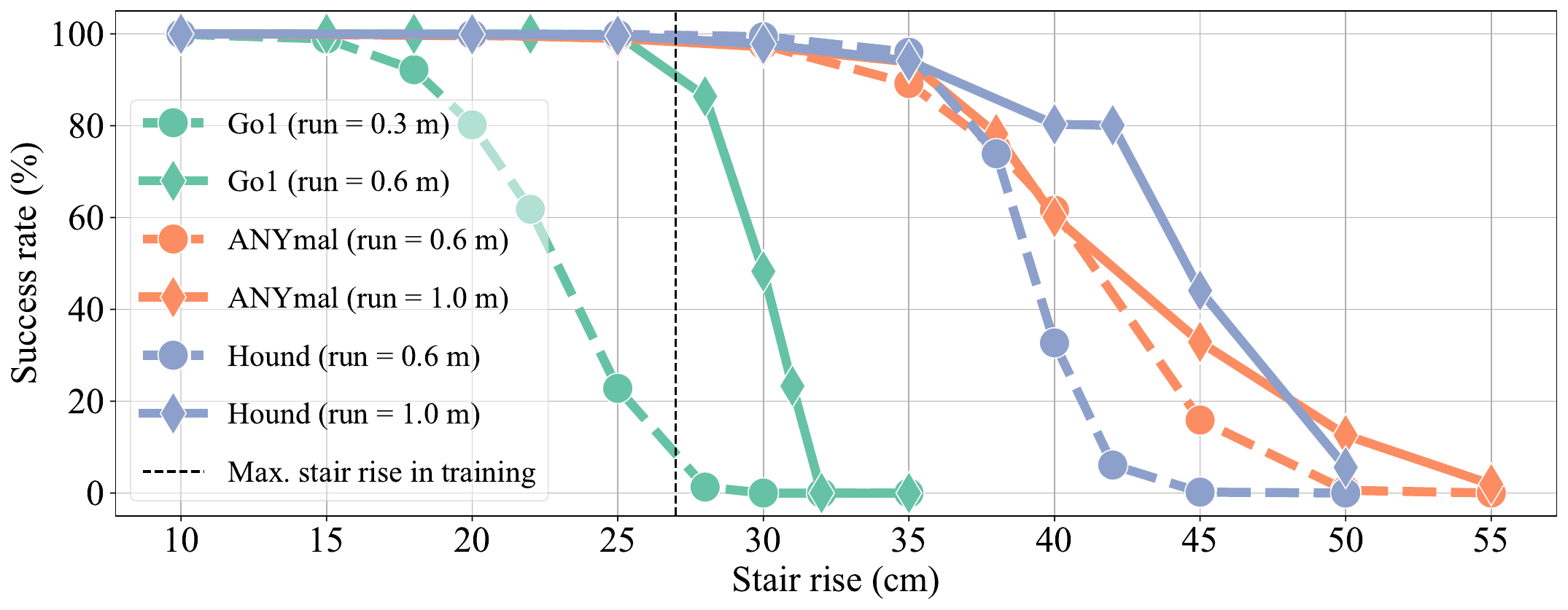}
	\end{subfigure}
	\caption{Success rates for climbing different obstacles using various quadrupedal robots. We trained DreamWaQ++ for Unitree Go1, ANYmal-C, and Hound. The maximum stair rise imposed during training for all robots is $27~\mathrm{cm}$.}
	\label{figure:supp_multi_robot}
\end{figure}

We evaluated the scalability of DreamWaQ++ by applying it to legged robots with different morphologies and sizes by training controllers for ANYmal-C~\cite{hutter2016anymal} and Hound~\cite{shin2022design} robots. Note that we used the same reward functions and its corresponding weights. Particularly, we adjusted only the robot-specific parameters such as the robot models, the motor operation limit, motor stiffness and damping parameters, desired foot clearance, desired body height, and maximum foot contact force.

\figinitref{figure:supp_multi_robot} presents the success rates of three different robots climbing stairs with varying runs and rises. For the Go1 robot, we used run values of $0.3~\mathrm{m}$ and $0.6~\mathrm{m}$, while for ANYmal-C and Hound, we used $0.6~\mathrm{m}$ and $1.0~\mathrm{m}$ runs to accommodate their longer trunk sizes. Success rates were measured from $1,\!000$ simulated robots, defined as the percentage reaching the last stair within $10~\mathrm{s}$.

As expected, Hound’s extensive joint operation range enabled it to traverse more challenging terrains than the Go1 and ANYmal-C, achieving an $80\%$ success rate on $42~\mathrm{cm}$-high stairs. Notably, the robots were only trained on obstacles up to $27~\mathrm{cm}$ in height, underscoring the strong adaptability of DreamWaQ++. The controller’s exploratory training maximizes hardware capability, with reward functions invariant to hardware differences.

This evaluation highlights DreamWaQ++'s versatility across various legged robots. Despite the sensitivity of deep RL algorithms to reward parameters, DreamWaQ++ demonstrated easy adaptability to different platforms without additional tuning.

    \subsection[Overcoming Large Obstacles]{Overcoming Large Obstacles\protect\mveightfoot}\label{supp:parkour}
\begin{figure}[t!]
	\centering 
	\begin{subfigure}[b]{0.48\textwidth}
		\includegraphics[width=1.0\textwidth]{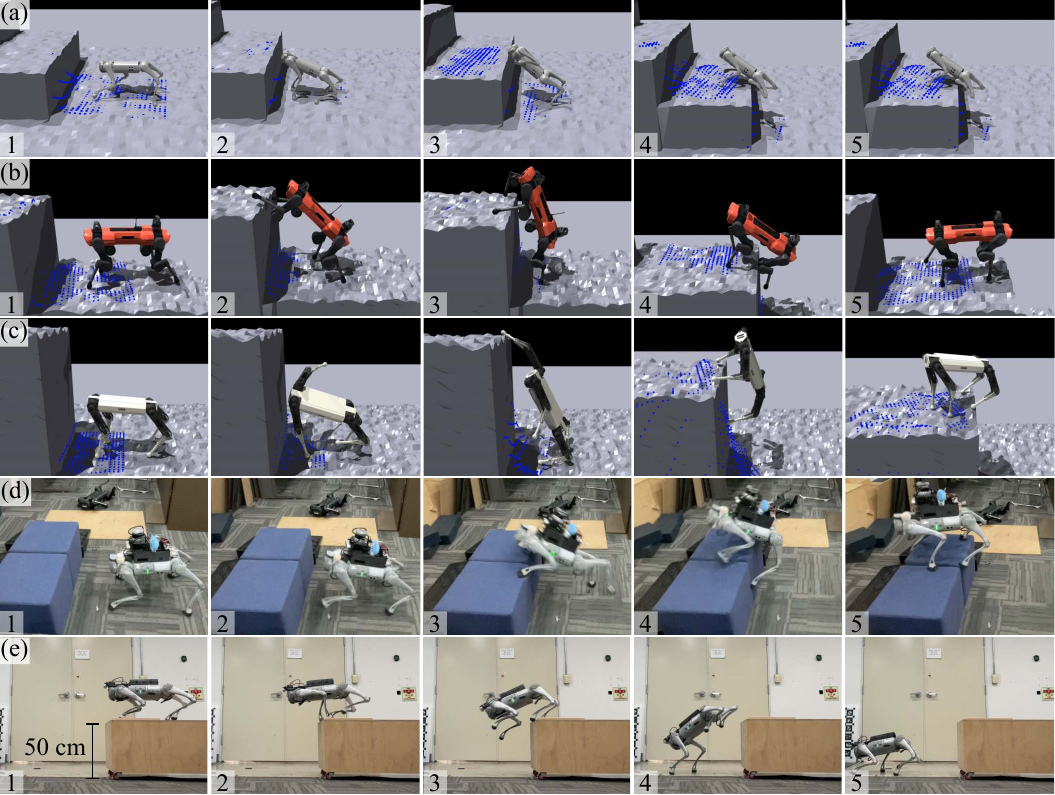}
	\end{subfigure}
	\caption{We trained DreamWaQ++ for (a)~Unitree Go1, (b)~ANYmal-C, and (c)~Hound to climb obstacles with a height of $0.6~\mathrm{m}$, $1.0~\mathrm{m}$, $1.5~\mathrm{m}$, respectively. (d)~A real world experiment was conducted using a Go1 robot with a $2.5~\mathrm{kg}$ payload on top of it.}
	\label{figure:supp_parkour_sim}
\end{figure}

Recently, there has been increasing interest in training quadrupedal robots for complex tasks such as jumping and leaping~\cite{hoeller2023anymal,zhuang2023robot}. These skills require robots to maximize actuator limits and perform highly agile motions. We further assessed the scalability of DreamWaQ++ for overcoming obstacles higher than the robot itself. We trained the Go1, ANYmal-C, and Hound robots in environments with extreme obstacle heights up to $0.6~\mathrm{m}$, $1.0~\mathrm{m}$, and $1.5~\mathrm{m}$, respectively. To enable learning of this agile skill, we made two modifications: (1) reducing the velocity tracking reward scale from $1.0$ to $0.1$, and (2) increasing the versatility gain scaling in the loss function from $0.1$ to $0.2$. Snapshots of the learned obstacle-climbing motions in simulation and real-world scenarios are shown in Fig.~\ref{figure:supp_parkour_sim}.

These modifications were complementary; without scaling up the versatility gain, reducing the velocity tracking reward alone led to a policy that resisted forward movement, while increasing versatility gain alone resulted in a conservative gait lacking skill discovery. Together, these adjustments allowed the policy to learn flexible gaits, yielding the agility required for complex tasks such as parkour.

Results in Fig.~\ref{figure:supp_parkour_sim} illustrate how controllers on different robots produce distinct motions for overcoming large obstacles. In Fig.~\ref{figure:supp_parkour_sim}(a), Go1 uses a jumping motion due to its small size. In contrast, ANYmal-C initially contacts the obstacle wall with its front legs (Fig.~\ref{figure:supp_parkour_sim}(b-2)), then uses them as anchors to climb (Fig.~\ref{figure:supp_parkour_sim}(b-3)). ANYmal-C’s large joint range and torque limits enable it to climb obstacles up to $1.5~\mathrm{m}$. Hound swings its right front leg widely to anchor on top, then propels itself upward with the rear legs (Fig.~\ref{figure:supp_parkour_sim}(c-4)).

In a real-world experiment shown in Fig.~\ref{figure:supp_parkour_sim}(d), we validated the sim-to-real robustness of the controller. A Go1 robot with an additional $2.5~\mathrm{kg}$ payload successfully climbed a $41~\mathrm{cm}$ obstacle, a soft sofa block representing a deformable surface not encountered in training.

In Fig.~\ref{figure:supp_parkour_sim}(e), the robot was placed on a $50~\mathrm{cm}$ stage, preventing simple terrain probing. Upon command, the robot leaped forward, avoiding rear leg collisions with the stage edges. This leap was facilitated by the velocity tracking relaxation, allowing the robot to pause at the edge (Fig.~\ref{figure:supp_parkour_sim}(e-2)) before initiating the leap. The robot kicks forward with its front legs, propels with the rear legs (Fig.~\ref{figure:supp_parkour_sim}(e-3)), and folds the rear legs to avoid collision (Fig.~\ref{figure:supp_parkour_sim}(e-4)). This demonstrates the versatility of the learned controller, serving as a prior adaptable to complex tasks.

\section{Ablation Study}\label{section:ablation}
\begin{table*}[ht!]
    \centering
    \caption{Ablation study on the impact of different architectures and regularization losses on the policy performance. The \protect\hlbest{best}, \protect\hlsecond{second best}, and \protect\hlworst{worst} performing methods for each test environment and metric are highlighted accordingly.}
    \label{table:ablation_encoding}
    \begin{tabular}{lccccccccc}
        \toprule
        \multicolumn{1}{c}{\multirow{2}{*}{\textbf{Method}}} 
        &\multicolumn{3}{c}{\textbf{Success rates}~$\textbf{[\%]}\uparrow$} 
        &\multicolumn{3}{c}{\textbf{Traveled distance}~$\textbf{[m]}\uparrow$} 
        &\multicolumn{3}{c}{\textbf{Calf collisions}~$\downarrow$} 
        \\ 
        
        &\multicolumn{1}{c}{Stairs-easy}
        &\multicolumn{1}{c}{Stairs-hard}
        &\multicolumn{1}{c}{Discrete}
        &\multicolumn{1}{c}{Stairs-easy}
        &\multicolumn{1}{c}{Stairs-hard}
        &\multicolumn{1}{c}{Discrete}
        &\multicolumn{1}{c}{Stairs-easy}
        &\multicolumn{1}{c}{Stairs-hard}
        &\multicolumn{1}{c}{Discrete} \\
        \hline\hline
        
        \textbf{Architecture} \\[0.1ex]
        \multicolumn{1}{l}{No memory}  
        & \alignedcell{white}{98.7} & \alignedcell{white}{85.1} & \alignedcell{second}{99.6} & \alignedcell{white}{8.4 \pm 1.2} & \alignedcell{white}{5.8 \pm 3.5} & \alignedcell{white}{11.5 \pm 0.9} & \alignedcell{worst}{30.1 \pm 0.8} & \alignedcell{worst}{35.2 \pm 10.2} & \alignedcell{worst}{27.4 \pm 20.1}  \\ 

        \multicolumn{1}{l}{Implicit memory}
        & \alignedcell{second}{99.1} & \alignedcell{white}{92.5} & \alignedcell{white}{99.5} & \alignedcell{white}{9.1 \pm 0.7} & \alignedcell{white}{8.8 \pm 1.3} & \alignedcell{white}{9.8 \pm 0.5} & \alignedcell{white}{24.1 \pm 5.1} & \alignedcell{white}{37.8 \pm 4.3} & \alignedcell{white}{15.9 \pm 2.7}  \\ 

        \multicolumn{1}{l}{No latent fusion}
        & \alignedcell{worst}{90.2} & \alignedcell{worst}{60.7} & \alignedcell{worst}{94.3} & \alignedcell{worst}{6.5 \pm 1.3} & \alignedcell{worst}{3.4 \pm 3.1} & \alignedcell{worst}{9.3 \pm 0.8} & \alignedcell{white}{22.3 \pm 2.1} & \alignedcell{white}{25.3 \pm 2.3} & \alignedcell{white}{13.5 \pm 2.6}  \\ 
        \midrule

        \textbf{Regularization} \\[0.1ex]
        \multicolumn{1}{l}{w/o contrastive loss}
        & \alignedcell{white}{98.4} & \alignedcell{second}{93.1} & \alignedcell{white}{99.2} & \alignedcell{second}{9.8 \pm 0.5} & \alignedcell{second}{8.9 \pm 1.4} & \alignedcell{white}{12.2 \pm 0.4} & \alignedcell{second}{12.3 \pm 1.8} & \alignedcell{second}{23.7 \pm 0.7} & \alignedcell{second}{8.4 \pm 0.9}  \\ 

        \multicolumn{1}{l}{w/o versatility gain}
        & \alignedcell{white}{98.2} & \alignedcell{white}{89.4} & \alignedcell{white}{99.3} & \alignedcell{white}{9.7 \pm 0.3} & \alignedcell{white}{8.8 \pm 1.5} & \alignedcell{second}{12.4 \pm 0.5} & \alignedcell{white}{15.2 \pm 0.5} & \alignedcell{white}{31.5 \pm 1.1} & \alignedcell{white}{11.3 \pm 1.4}  \\ 
        \midrule

        \multicolumn{1}{l}{DreamWaQ++}
        & \alignedcell{best}{99.5} & \alignedcell{best}{97.8} & \alignedcell{best}{99.8} & \alignedcell{best}{10.3 \pm 0.8} & \alignedcell{best}{9.5 \pm 1.2} & \alignedcell{best}{12.5 \pm 0.1} & \alignedcell{best}{10.5 \pm 0.4} & \alignedcell{best}{18.3 \pm 2.5} & \alignedcell{best}{5.8 \pm 0.2}  \\ 
        \bottomrule
    \end{tabular}
\end{table*}

\subsection{Quantitative Analysis on Policy Performance}
We evaluated the effectiveness of various multi-modal encoding strategies for learning a context representation that captures the dynamics of a quadrupedal robot. This ablation study examines the impact of various encoding methods on both policy performance and training dynamics. We trained several variants of the multi-modal context encoder, each using a different strategy as follows:

\begin{itemize}
    \item \textbf{No memory.} This variant does not use any memory structure; the encoder processes only the current proprioceptive and exteroceptive measurements.
    
    \item \textbf{Implicit memory.} This variant employs an LSTM to implicitly encode the history of proprioceptive and exteroceptive inputs without explicitly storing past information.
    
    \item \textbf{No latent fusion.} In this variant, proprioceptive and exteroceptive features are simply concatenated, with no dedicated fusion mechanism—treating them as separate input modalities.
    
    
    \item \textbf{DreamWaQ++ w/o contrastive loss.} This variant trains DreamWaQ++ without the contrastive loss which is used to align the latent representations of proprioceptive and exteroceptive features.
    
    \item \textbf{DreamWaQ++ w/o versatility gain.} This variant trains DreamWaQ++ without the versatility gain which encourages the policy to explore a wider range of actions and behaviors.
\end{itemize}

We evaluated all variants in a simulated environment using $1,\!000$ robots; each of the robots commanded to walk forward for $20~\textrm{s}$. The evaluations were conducted in three test environments: 
\begin{itemize}
    \item \textbf{Stairs-easy}. Continuous stairs with a step height of $10~\textrm{cm}$.
    \item \textbf{Stairs-hard}. Continuous stairs with a step height of $20~\textrm{cm}$.
    \item \textbf{Discrete}. Discrete obstacle terrain with obstacle heights ranging from $5~\textrm{cm}$ to $25~\textrm{cm}$.
\end{itemize}

For each variant, we report the average success rate, the average distance traveled, and the total number of calf collisions across the test environments. The results are summarized in Table~\ref{table:ablation_encoding}.

The results presented in Table~\ref{table:ablation_encoding} validate that DreamWaQ++ consistently outperforms all other variants across all test environments and evaluation metrics. Notably, we observed that the latent fusion mechanism plays a critical role: the variant without latent fusion exhibited the lowest success rates and shortest traveled distances despite not having the highest number of calf collisions. This suggests that latent fusion is crucial for maintaining robustness and stability, especially in situations where proprioceptive and exteroceptive inputs are inconsistent.

We also observed that the memory plays a significant role in collision avoidance. The variant without any memory structure resulted in the highest number of calf collisions across all environments. While the implicit memory variant performed worse than DreamWaQ++, it still achieved a reasonable level of performance, suggesting that even approximate temporal encoding can help improve safety.

Furthermore, in terms of regularization, the ablation study highlights the importance of both the contrastive loss and the versatility gain in DreamWaQ++. Removing the contrastive loss led to modest declines in all metrics, underscoring its value in aligning the latent spaces of proprioceptive and exteroceptive inputs, thereby improving policy learning. In contrast, removing the versatility gain resulted in more substantial performance degradation. This component encourages the policy to explore a broader range of behaviors. Without it, the policy tends to converge to a narrower action distribution, reducing adaptability and robustness in diverse terrains.

\subsection{Embedding Analysis}
    \begin{figure}[t!]
        \centering 
        \begin{subfigure}[b]{0.48\textwidth}
            \includegraphics[width=1.0\textwidth]{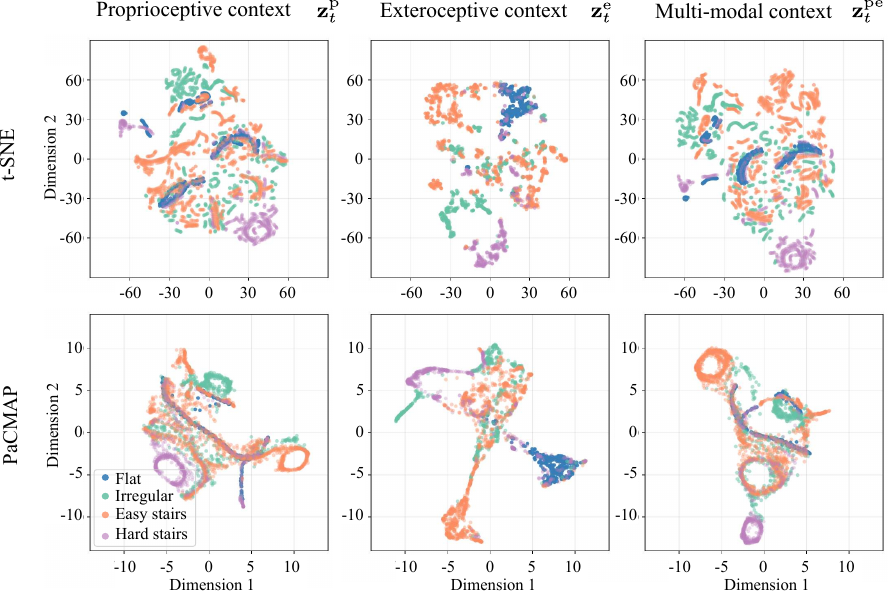}
        \end{subfigure}
        \caption{~Embedding visualization of the multi-modal context encoded by the proposed context encoder in different environments using PacMAP~\cite{wang2021understanding}. The highly disentangled multi-modal context serves as an informative prior for informing about the environment to the policy.}
        \label{figure:ablation_embedding}
    \end{figure}

    The visualized embeddings in Fig.~\ref{figure:ablation_embedding} reveal a distinctive ellipsoidal pattern in the proprioceptive context $\textbf{z}^\mathrm{p}_t$, attributed to the dynamic motion of the robot's feet, as $\textbf{z}^\mathrm{p}_t$ integrates various proprioceptive measurements. Notably, the ellipsoid’s size diminishes on more challenging terrains, likely reflecting the robot’s foot swing period. The controller adjusts by orchestrating quicker foot swings over difficult terrains, ensuring frequent ground contact to enhance locomotion stability.

    In contrast, the exteroceptive context $\textbf{z}^\mathrm{e}_t$ displays clearer inter-class separation among environments compared to $\textbf{z}^\mathrm{p}_t$. However, similarities among some embeddings likely result from simplified exteroceptive input, primarily 3D voxels ahead of the robot. The exteroceptive encoder effectively captures relevant geometric features, filtering out unnecessary details from raw 3D points.

    For the multi-modal context $\textbf{z}^\mathrm{pe}_t$, which fuses proprioceptive and exteroceptive inputs, Fig.~\ref{figure:ablation_embedding} displays distinct clusters based on terrain difficulty. Embeddings from \texttt{flat}, \texttt{easy stairs}, and \texttt{irregular} terrains are partially clustered near the origin due to similarities in obstacle height, despite differences in obstacle placement and density. However, significant disentanglement is evident within the \texttt{easy stairs} and \texttt{irregular} terrain clusters, facilitated by proprioceptive information capturing terrain properties under the robot.

    Simultaneously, the circular pattern from $\textbf{z}^\mathrm{p}_t$ persists in $\textbf{z}^\mathrm{pe}_t$, aiding in the correction of unreliable exteroceptive data. This feature contributes to the clear separation between \texttt{easy stairs} and \texttt{irregular} terrains, underscoring exteroception's auxiliary role in adapting the robot's gait to avoid upcoming obstacles.

\subsection{Latent Modulation for Gait Control}
    \begin{figure}[t!]
        \centering 
        \begin{subfigure}[b]{0.48\textwidth}
            \includegraphics[width=1.0\textwidth]{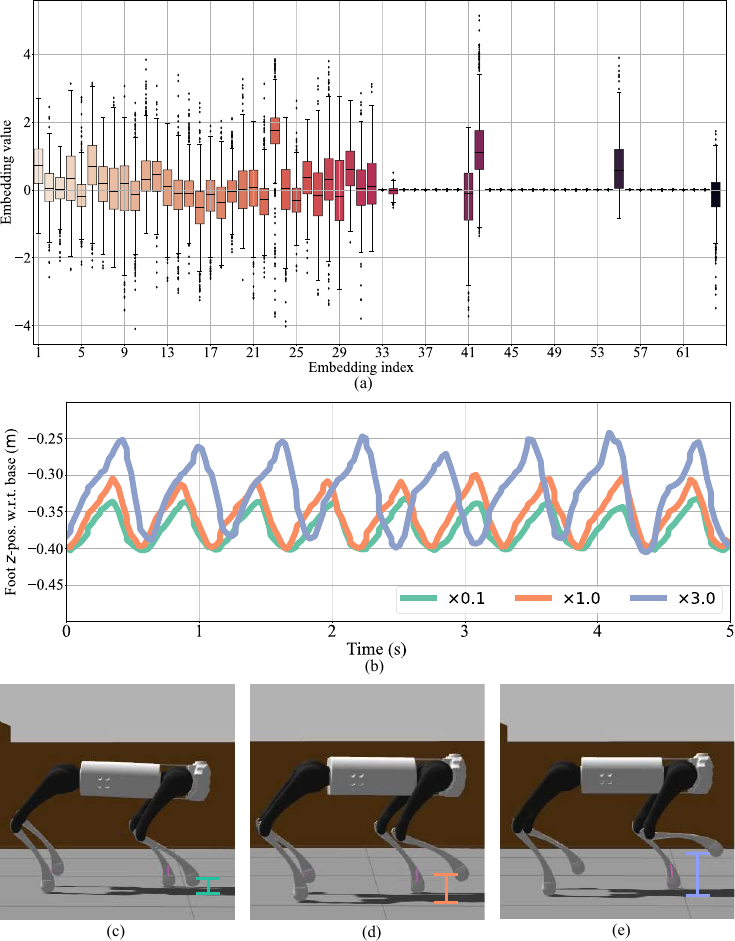}
        \end{subfigure}
        \caption{(a)~Boxplots of the multi-modal contexts in an irregular terrain, showing the distribution of embeddings activation from the multi-modal context and highlighting the contrast between activations in the exteroceptive context. Latent modulation on some exteroceptive embedding proportionally affects the gait height. (b)~The robot exhibits different locomotion styles when particular latent variables ($41$, $42$, $55$, and $64$th embeddings) were scaled with (c)~$0.1$, (d)~$1.0$, and (e)~$3.0$ times of its original value.}
        \label{figure:ablation_gait}
    \end{figure}

    Fig.~\ref{figure:ablation_gait} illustrates the distribution of each latent feature as the robot traversed irregular terrains. Embedding indices from \texttt{1} to \texttt{32} correspond to $\textbf{z}^\mathrm{p}_t$, and indices from \texttt{33} to \texttt{64} correspond to $\textbf{z}^\mathrm{e}_t$. While $\textbf{z}^\mathrm{p}_t$ features display a uniform distribution, $\textbf{z}^\mathrm{e}_t$ contains four embeddings with distinct scaling differences compared to other exteroceptive features. This finding raises the question: \textit{Do these four exteroceptive embedding features correlate with the robot’s foot swing motion?}

    To address this question, we conducted an experiment by modulating these four $\textbf{z}^\mathrm{e}_t$ features. The results in Fig.~\ref{figure:ablation_gait}(b) indicate a clear trend: \textit{scaling up the latent feature decreases gait frequency while increasing gait height, and vice versa}. The resulting gait pattern resembles that used in stair-climbing scenarios, suggesting that the multi-modal context encoder effectively activates key latent variables that directly influence gait. A limitation of this approach is that the activated latent variables may vary across training seeds, as they are learned in an unsupervised manner with multiple randomizations. However, upon convergence, these critical latent features reliably emerge within the latent space.


\subsection{Cross-Modal Feature Correlation}
    \begin{figure}[t!]
        \centering 
        \begin{subfigure}[b]{0.48\textwidth}
            \includegraphics[width=1.0\textwidth]{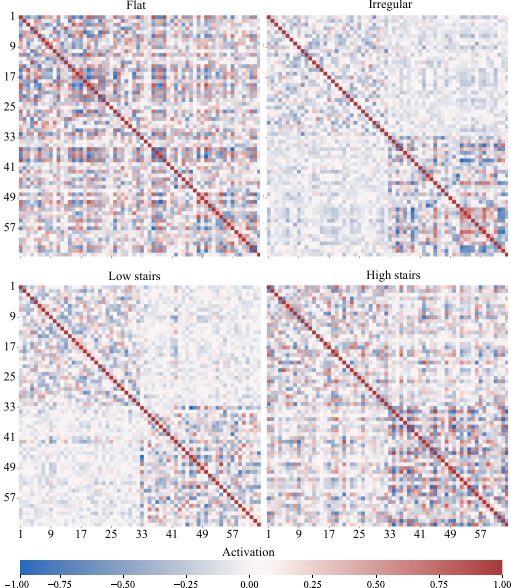}
        \end{subfigure}
        \caption{Cross-modal feature correlations between proprioceptive and exteroceptive context vectors. The heatmap shows the correlation between the embedding features, where low correlation indicates substantial mismatches between proprioceptive and exteroceptive information.}
        \label{figure:ablation_heatmap}
    \end{figure}

    Cross-modal correlations between context vectors are visualized as heatmap plots in Fig.~\ref{figure:ablation_heatmap}, obtained by computing cross-correlations between embedding features. Low cross-modal correlation on irregular terrains indicates substantial mismatches between proprioceptive and exteroceptive information. Additionally, stronger correlations are observed within the exteroceptive data due to its direct observability, as opposed to proprioception, which can only approximate the terrain beneath the robot. In contrast, higher cross-modal correlations on flat terrain result from the similar predictions of terrain structure provided by both proprioceptive and exteroceptive inputs.

    Additionally, it can be observed that cross-modal correlation is higher on high stairs than on low stairs. This is because, on high stairs, the robot’s feet are more likely to be in contact with the surface, providing more accurate proprioceptive information. In contrast, on low stairs, the robot’s feet are more often in the air, leading to less accurate proprioceptive feedback. This discrepancy between proprioceptive and exteroceptive information results in lower cross-modal correlation on low stairs compared to high stairs.

\subsection{Locomotion under Exteroception Failure}\label{supp:exteroception_failure}
    \begin{figure}[t!]
        \centering 
        \begin{subfigure}[b]{0.48\textwidth}
            \includegraphics[width=1.0\textwidth]{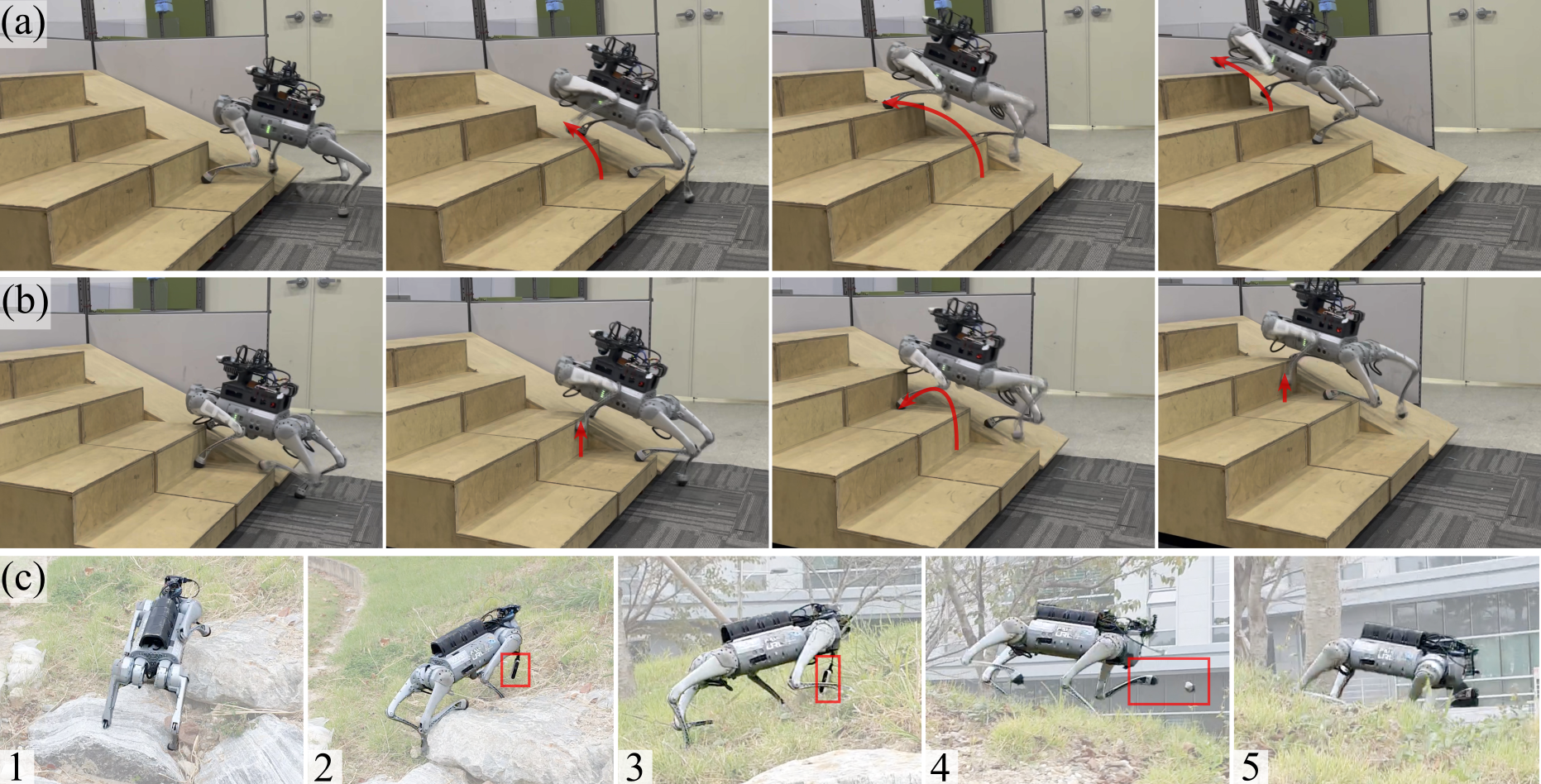}
        \end{subfigure}
        \caption{Locomotion under exteroception failures. (a)~The robot swiftly adapts its gait when climbing the stairs under normal exteroception condition. Meanwhile, (b)~the robot's foot collides with the stairs and yield a foot-trapping reflex when white noise is provided as exteroception input. (c)~In an extreme failure case, the robot adapts to the detached camera by making contact with the ground using its feet and knees to ensure a stable pose.}
        \label{figure:supp_stairs_swing}
    \end{figure}

    Fig.~\ref{figure:supp_stairs_swing} shows the emergent behaviors of DreamWaQ++ when climbing stairs. This experiment ablates foot swing adaptation by providing the policy with a white noise input. The robot is commanded to climb the stairs under both conditions. The red arrows in Fig.~\ref{figure:supp_stairs_swing}(a) illustrate the foot swing motion of the robot when exteroception functions normally, allowing the robot to adapt its foot swing trajectory to climb two stairs at once. In contrast, when the robot receives white noise input, as in Fig.~\ref{figure:supp_stairs_swing}(b), the robot's foot tends to collide with the stairs. However, the robot adapts with a foot-trapping reflex, dragging its foot along the stair's vertical surface before placing it on the next step.

    Additionally, the experiment in Fig.~\ref{figure:supp_stairs_swing}(c) shows the robot climbing large rocks by swinging its feet extensively, resulting in strong vibrations. These vibrations eventually caused the camera to fall, yielding depth point cloud measurements with significant calibration error. The problem worsens when the camera becomes completely detached from the robot (see Fig.~\ref{figure:supp_stairs_swing}(c-4)). Under these conditions, the controller no longer receives new data streams. Interestingly, the robot adapts its gait to move by making contact with the ground using its feet and knees (Fig.~\ref{figure:supp_stairs_swing}(c-5)). This motion results in a more stable pose, with the robot adapting additional ground contacts to handle high-risk locomotion when exteroception becomes extremely unreliable.

\subsection{Terrain Reconstruction}\label{supp:recons}
\begin{figure}[t!]
        \centering 
        \begin{subfigure}[b]{0.48\textwidth}
            \includegraphics[width=1.0\textwidth]{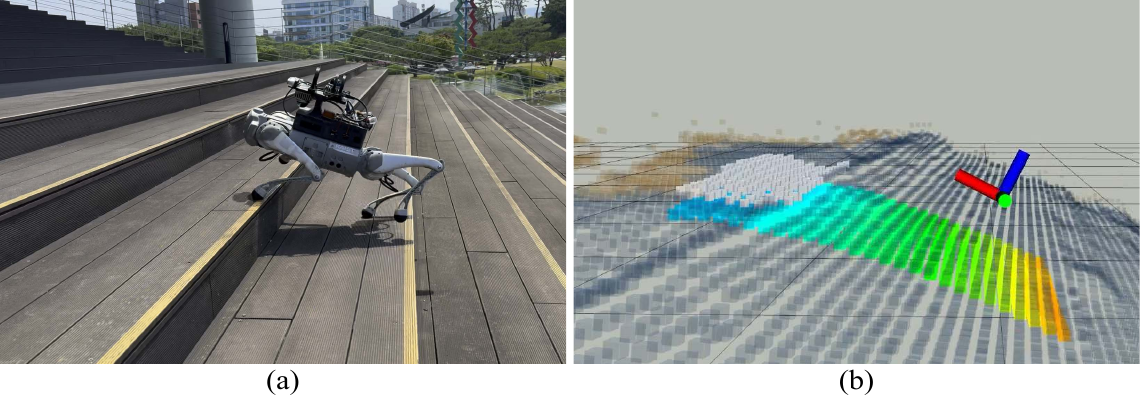}
        \end{subfigure}
        \caption{(a)~An snapshot of the scene during the asynchronous stair race using DreamWaQ++. (b)~The terrain map is reconstructed from the recorded latent features and the ground truth terrain map is constructed using~\cite{oh2024trip}. The white points are the forward 3D scan input, while the reconstruction points surrounding the robots are colored based on the height relative to the robot's base.}
        \label{figure:supp_recons}
    \end{figure}

    We decoded the latent features from the context encoder recorded during the asynchronous stair race experiment to reconstruct the terrain map, as shown in Fig.~\ref{figure:supp_recons}. The reconstructed terrain map resembles the ground truth terrain map constructed with~\cite{oh2024trip}.

    However, the reconstructed map is less accurate than the ground truth due to three factors:
    \begin{enumerate}
        \item The robot’s exteroception is limited to front-facing 3D points, and the encoder’s memory is not long-term.
        \item The encoder-decoder structure lacks residual connections, as in~\cite{hoeller2023anymal}, focusing the latent representation on relevant features.
        \item Variational autoencoder regularization in the latent space prioritizes feature disentanglement over precise reconstruction.
    \end{enumerate}
    Nonetheless, this slight reduction in reconstruction accuracy is acceptable, as the primary goal of the context encoder is to learn a structured representation for control rather than to fully reconstruct the terrain map.

\begin{figure}[t!]
        \centering 
        \begin{subfigure}[b]{0.48\textwidth}
            \includegraphics[width=1.0\textwidth]{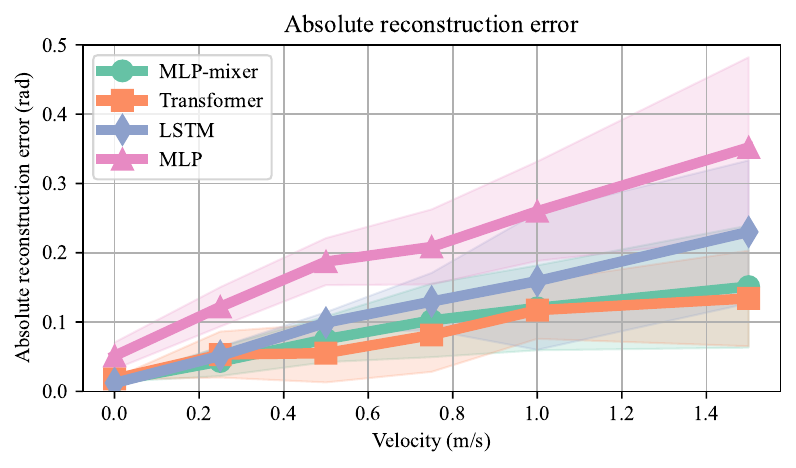}
        \end{subfigure}
        \caption{Future state prediction error using different models and varying robot velocities. The MLP-mixer model used in DreamWaQ++ shows comparable performance to the Transformer model while being more lightweight and computationally efficient.}
        \label{figure:supp_nextstate}
    \end{figure}

    We further evaluated the impact of the backbone sequence model on the controller's performance by comparing the controller’s performance across different sequence models. Performance was assessed by measuring the accuracy of future joint position predictions, as shown in Fig.~\ref{figure:supp_nextstate}.

    The results in Fig.~\ref{figure:supp_nextstate} show that the proposed MLP-mixer architecture performs comparably to the Transformer model. The MLP-mixer offers the advantage of being more lightweight and computationally efficient than the Transformer. As indicated by the curve, the prediction error increases with the robot's velocity due to the more dynamic nature of the motion, which requires higher accuracy in future state predictions.

\subsection{State Estimation Accuracy}
We evaluate the proprioceptive-based state estimation accuracy of DreamWaQ++ which utilizes an MLP-mixer architecture, and compare it with other common approaches such as fully connected networks~\cite{nahrendra2023dreamwaq} and RNN-based architectures~\cite{ji2022concurrent}. Since accurate state estimation is crucial for constructing the hierarchical exteroceptive memory, we also evaluate how different state estimation architectures impact the reconstruction quality for the exteroceptive input.

We conducted a controlled experiment in a simulated stair environment, where the robot traverses a staircase with step heights uniformly sampled from the range $[0.1,~0.3]~\textrm{m}$. To isolate perception performance, the robot was commanded to move forward at a constant speed of $0.5~\textrm{m/s}$. Subsequently, we measured the absolute reconstruction error across $187$ height points in the robot's surrounding area. The results are summarized in Table~\ref{table:state_estimation_accuracy}. All measurements represent the mean over $5~\textrm{s}$ of continuous walking using $1,\!000$ simulated robots.

\begin{table}[ht!]
    \centering
    \caption{State estimation and terrain reconstruction accuracy.}
    \label{table:state_estimation_accuracy}
    \begin{tabular}{lcccc}
        \toprule
        \multicolumn{1}{c}{\multirow{2}{*}{\textbf{Architecture}}} 
        &\multicolumn{3}{c}{\textbf{Estimation error} $\downarrow$}
        &\multicolumn{1}{c}{\textbf{Reconstruction}} \\

        &\multicolumn{1}{c}{$v_x~\textrm{[m/s]}$}
        &\multicolumn{1}{c}{$v_y~\textrm{[m/s]}$}
        &\multicolumn{1}{c}{$v_z~\textrm{[m/s]}$}
        &\multicolumn{1}{c}{\textbf{error}~$\textrm{[m]}$ $\downarrow$} \\
        \midrule
        
        MLP         & $0.0571$ & $0.0739$ & $0.0715$ & $0.1978$ \\
        RNN         & $0.0497$ & $0.0613$ & $0.0514$ & $0.1253$ \\
        MLP-mixer   & $0.0363$ & $0.0331$ & $0.0385$ & $0.0674$ \\
        \bottomrule
    \end{tabular}
\end{table}

The results indicate that the MLP-mixer architecture outperforms both the fully connected MLP and RNN-based architectures in terms of state estimation accuracy, achieving the lowest error across all velocity components. This superior performance is attributed to the MLP-mixer's ability to effectively capture temporal dependencies and interactions among different proprioceptive modalities, resulting in a more accurate representation of the robot's state. Furthermore, the improved state estimation contributes to a more precise reconstruction of the surrounding terrain, as shown by the lowest reconstruction error among the evaluated architectures.

\section{Conclusion}
    We have proposed DreamWaQ++, a resilient yet lightweight obstacle-aware locomotion controller that enhances the resilience of its precursor, DreamWaQ~\cite{nahrendra2023dreamwaq}. DreamWaQ++ reduces the need for expensive onboard computation and increases versatility by incorporating 3D point cloud data as an exteroceptive modality. Our experiments demonstrate that the proposed controller exhibits enhanced explainability, opening possibilities for integration with model-based counterparts or higher-level planning modules to enable greater autonomy. A promising avenue for future work involves integrating an active tilting mechanism into the camera mount using an additional servo motor. By simultaneously learning both locomotion and camera tilting, this approach could yield a controller that actively maximizes its observability, resembling the active sensing behavior of animals.

\bibliographystyle{IEEEtran}
\bibliography{./main,./utils/IEEEabrv}

\begin{IEEEbiography}[{\includegraphics[width=1in,height=1.25in]{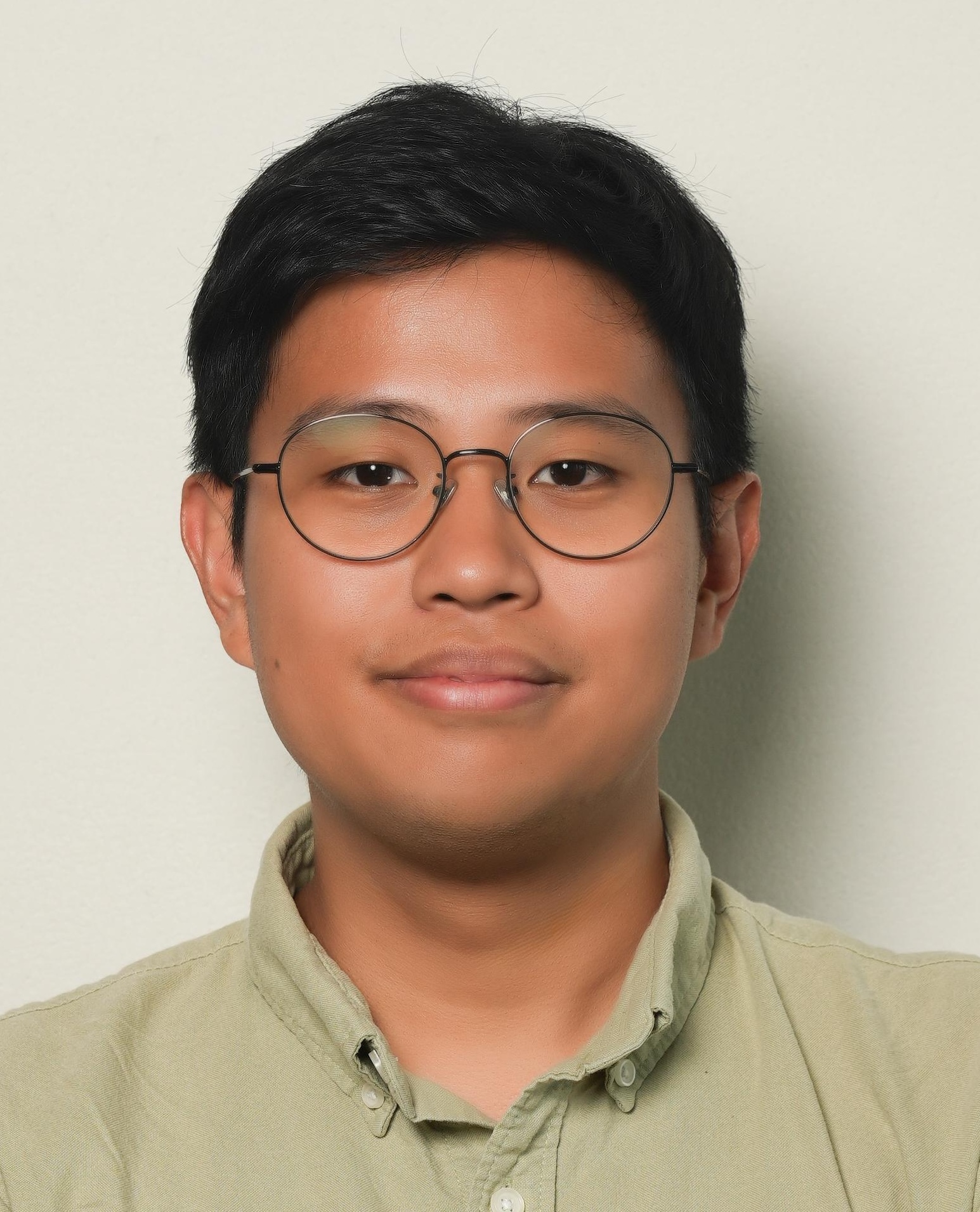}}]{I Made Aswin Nahrendra}
    received the B.S. and M.S. degrees in electrical engineering from Bandung Institute of Technology, Bandung, Indonesia, in 2018 and 2019, respectively. He received the Ph.D. degree in robotics program and electrical engineering from Korea Advanced Institute of Science and Technology (KAIST), Daejeon, Korea, in 2024. He was a postdoctoral fellow in the Information \& Electronics Research Institute, KAIST, Daejeon, Republic of Korea from 2024 to 2025. He is currently a research scientist in the Physical AI Team, AI Research Center at KRAFTON. His research interests include reinforcement learning, control, and robot learning for legged robots.
\end{IEEEbiography}
\vskip -2\baselineskip plus -1fil

\begin{IEEEbiography}[{\includegraphics[width=1in,height=1.25in]{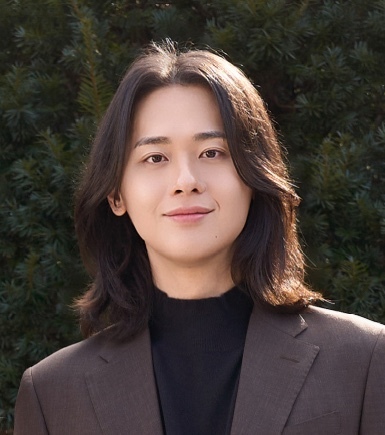}}]{Byeongho Yu}
	received the M.S. and Ph.D. degrees in Electrical Engineering from Korea Advanced Institute of Science and Technology (KAIST), Daejeon, Korea, in 2021 and 2025, respectively. He is currently the CEO of URobotics. His research interests are primarily in the field robotics, including visual-inertial-leg odometry and state estimation for legged robots, with a particular focus on robust autonomy in unstructured environments.
\end{IEEEbiography}
\vskip -2\baselineskip plus -1fil

\begin{IEEEbiography}[{\includegraphics[width=1in,height=1.25in]{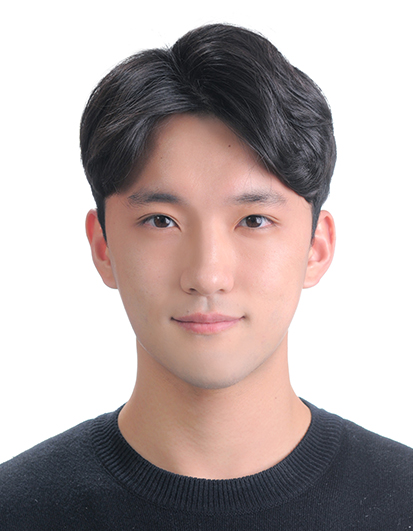}}]{Minho Oh}
	received the B.S. degree in convergence engineering from Daegu Gyeongbuk Institute of Science and Technology (DGIST), Daegu, Korea, in 2020, and the M.S. degree in electrical engineering from the Korea Advanced Institute of Science and Technology (KAIST), Daejeon, Korea, in 2021. He is currently pursuing the Ph.D. degree in electrical engineering, KAIST, and leads the spatial intelligence team as the CTO of URobotics. His research interests include SLAM and terrain traversability mapping for enhanced autonomous navigation.
\end{IEEEbiography}
\vskip -2\baselineskip plus -1fil

\begin{IEEEbiography}[{\includegraphics[width=1in,height=1.25in]{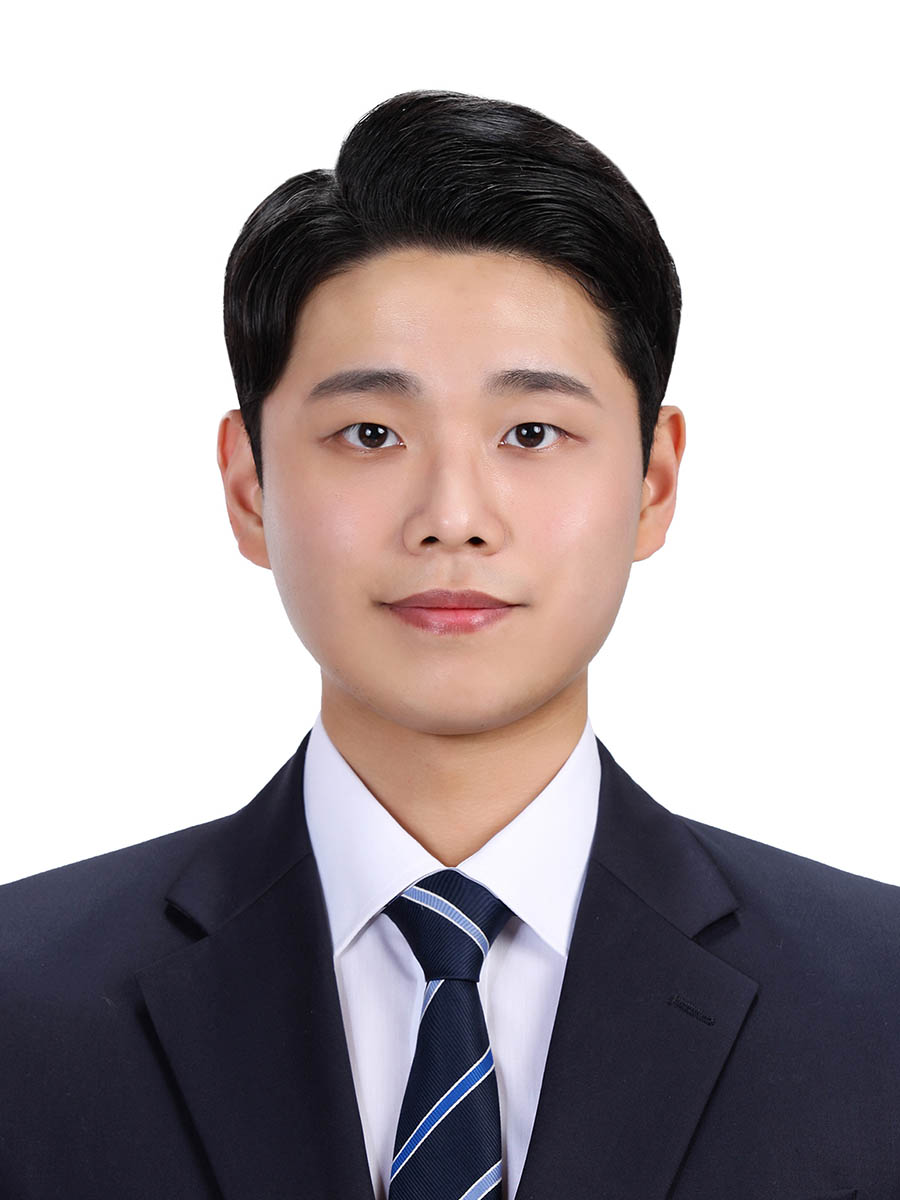}}]{Dongkyu Lee}
	received the B.S. degree in Electrical and Computer Engineering from the University of Seoul, Seoul, Korea, in 2021, and the M.S. degree in the Robotics Program from the Korea Advanced Institute of Science and Technology (KAIST), Daejeon, Korea, in 2023. He is currently pursuing the Ph.D. degree in Electrical Engineering at KAIST and is currently the CTO of URobotics Corp. His research interests include robot navigation, path planning, decision-making, and robot learning.
\end{IEEEbiography}
\vskip -2\baselineskip plus -1fil

\begin{IEEEbiography}[{\includegraphics[width=1in,height=1.25in]{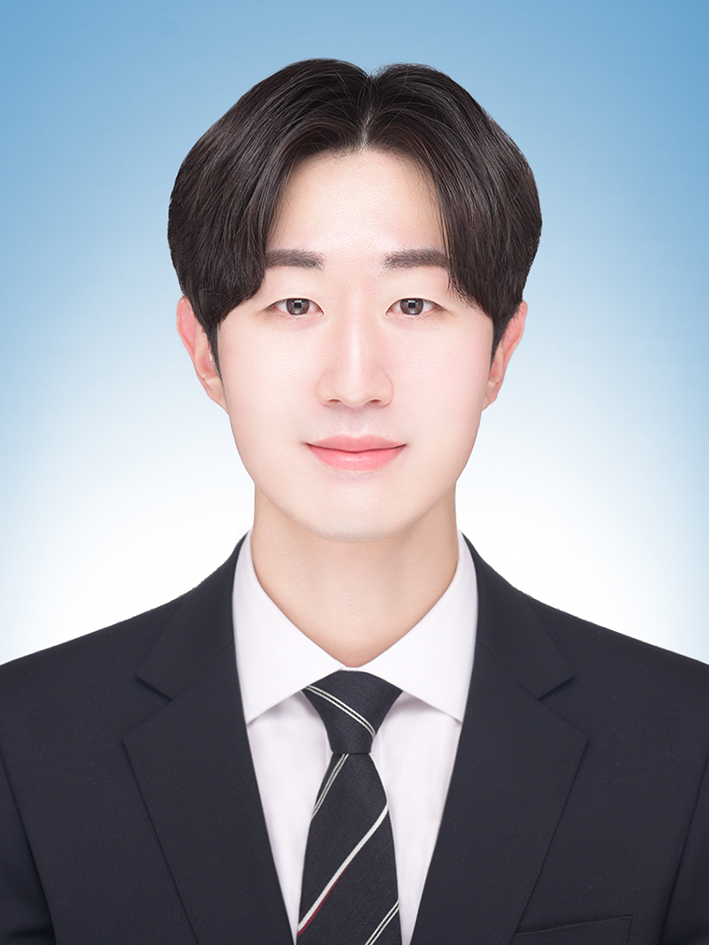}}]{Seunghyun Lee}
	received the B.S. and M.S. degrees in the school of electrical engineering from Korea Advanced Institute of Science and Technology (KAIST), Daejeon, Republic of Korea, in 2022 and 2024, respectively. He is currently pursuing the Ph.D. degree at Urban Robotics Lab in the school of electrical engineering, KAIST. His research interests include legged robot locomotion and deep reinforcement learning.
\end{IEEEbiography}
\vskip -2\baselineskip plus -1fil

\begin{IEEEbiography}[{\includegraphics[width=1in,height=1.25in]{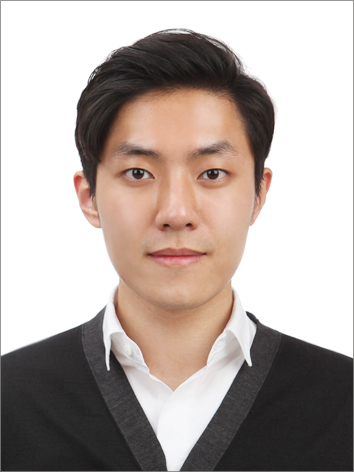}}]{Hyeonwoo Lee}
	received the B.S. and M.S. degrees in mechanical engineering from Yonsei University, Seoul, Republic of Korea, in 2020 and 2022, respectively. He is currently pursuing the Ph.D. degree in electrical engineering from the Korea Advanced Institute of Science and Technology with the Urban Robotics Lab. His research interests include reinforcement learning, motion and path planning for quadruped robots, and visual navigation.
\end{IEEEbiography}
\vskip -2\baselineskip plus -1fil

\begin{IEEEbiography}[{\includegraphics[width=1in,height=1.0in]{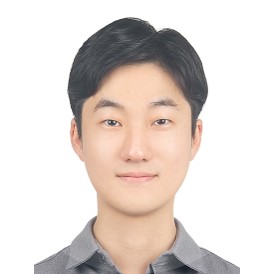}}]{Hyungtae Lim}
	received the B.S. degree in mechanical engineering, and M.S. and Ph.D. degrees in electrical engineering from the Korea Advanced Institute of Science and Technology (KAIST), Daejeon, Republic of Korea, in 2018, 2020, and 2023, respectively. He was a postdoctoral fellow in the Information \& Electronics Research Institute, KAIST, Daejeon, Republic of Korea from 2023 to 2024. He is currently a postdoctoral associate in the Laboratory for Information \& Decision Systems~(LIDS), Massachusetts Institute of Technology~(MIT), Cambridge, MA, USA.
    His research interests include SLAM (simultaneous localization and mapping), 3D registration, 3D perception, long-term map management, deep learning, and spatial AI.
\end{IEEEbiography}
\vskip -2\baselineskip plus -1fil

\begin{IEEEbiography}[{\includegraphics[width=1in,height=1.25in]{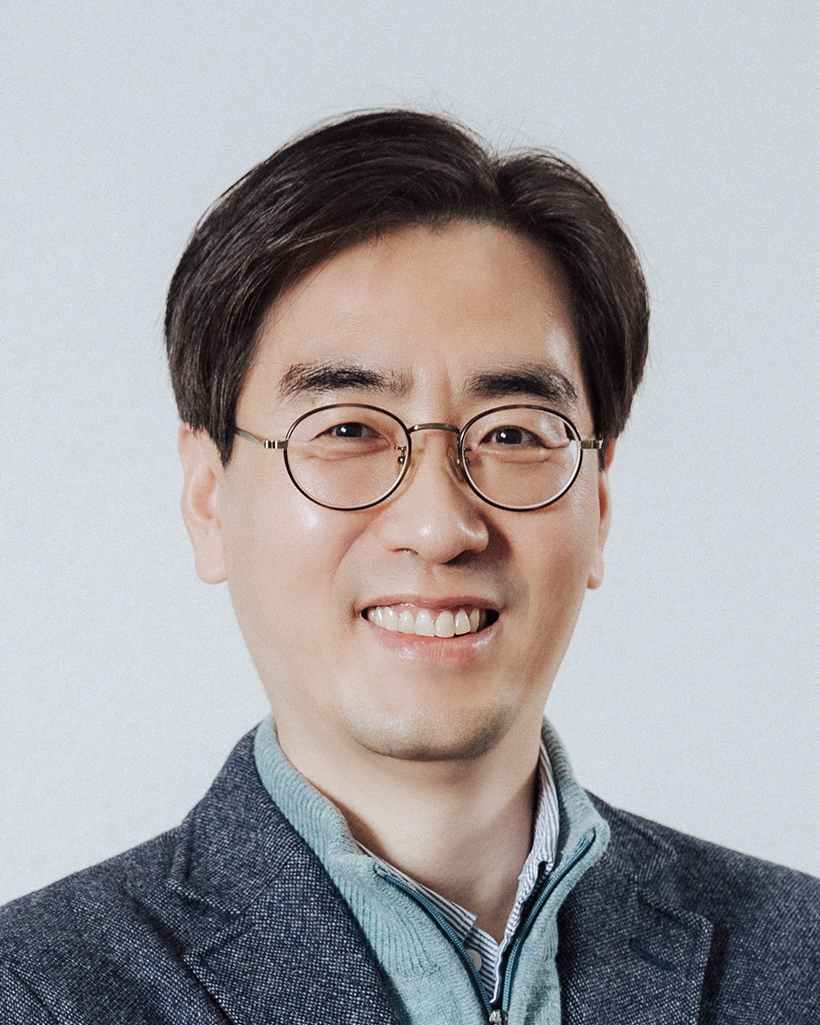}}]{Hyun Myung}
	received the B.S., M.S., and Ph.D. degrees in electrical engineering from the Korea Advanced Institute of Science and Technology (KAIST), Daejeon, Korea, in 1992, 1994, and 1998, respectively. He was a Senior Researcher with the Electronics and Telecommunications Research Institute, Daejeon, from 1998 to 2002, a CTO and the Director with the Digital Contents Research Laboratory, Emersys Corporation, Daejeon, from 2002 to 2003, and a Principle Researcher with the Samsung Advanced Institute of Technology, Yongin, Korea, from 2003 to 2008. Since 2008, he has been a Professor with the Department of Civil and Environmental Engineering, KAIST, and he was the Chief of the KAIST Robotics Program. From 2019, he is a Professor with the School of Electrical Engineering. His current research interests include autonomous robot navigation, SLAM (simultaneous localization and mapping), spatial AI/ML, and swarm robots.
\end{IEEEbiography}

\end{document}